\documentclass{article} %
\usepackage{iclr2021_conference,times}

\usepackage{amsmath,amsfonts,bm}

\def\eqref#1{equation~\ref{#1}}

\def\1{\bm{1}}

\def\vx{{\bm{x}}}
\def\vy{{\bm{y}}}
\def\vz{{\bm{z}}}

\def\mI{{\bm{I}}}

\DeclareMathAlphabet{\mathsfit}{\encodingdefault}{\sfdefault}{m}{sl}
\SetMathAlphabet{\mathsfit}{bold}{\encodingdefault}{\sfdefault}{bx}{n}

\def\gC{{\mathcal{C}}}

\def\gN{{\mathcal{N}}}

\def\gT{{\mathcal{T}}}

\newcommand{\E}{\mathbb{E}}

\newcommand{\R}{\mathbb{R}}

\newcommand{\KL}{D_{\mathrm{KL}}}

\usepackage{hyperref}
\usepackage{url}
\usepackage{graphicx} %
\usepackage{caption,subcaption} %
 \usepackage{todonotes}
\usepackage[capitalise]{cleveref}
\usepackage{nicefrac}
\usepackage{array}
\usepackage{wrapfig}
\usepackage{placeins}

\title{Uncertainty in Neural Processes}

\author{Saeid Naderiparizi$^{1}$, Kenny Chiu$^{2}$, Benjamin Bloem-Reddy$^{2}$, Frank Wood$^{1,3,4}$ \\
$^{1}$Department of Computer Science, University of British Columbia\\
$^{2}$Department of Statistics, University of British Columbia\\
$^{3}$MILA\\
$^{4}$CIFAR AI Chair\\
\texttt{\{saeidnp, fwood\}@cs.ubc.ca}, \texttt{\{kenny.chiu, benbr\}@stat.ubc.ca}
}

\renewcommand{\paragraph}[1]{\textbf{#1}\quad}

\begin{document}

\maketitle

\begin{abstract}
    We explore the effects of architecture and training objective choice on amortized posterior predictive inference in probabilistic conditional generative models.  We aim this work to be a counterpoint to a recent trend in the literature that stresses achieving good samples when the amount of conditioning data is large.  We instead focus our attention on the case where the amount of conditioning data is small.  We highlight specific architecture and objective choices that we find lead to qualitative and quantitative improvement to posterior inference in this low data regime.  Specifically we explore the effects of choices of pooling operator and variational family on posterior quality in neural processes.  %
    Superior posterior predictive samples drawn from our novel neural process architectures are demonstrated via image completion/in-painting experiments.
\end{abstract}

\section{Introduction}
What makes a probabilistic conditional generative model {\em good}?  The belief that a generative model is good if it produces samples that are indistinguishable from those that it was trained on \citep{hinton2007recognize} is widely accepted, and understandably so.
This belief also applies when the generator is conditional, though the standard becomes higher: conditional samples must be indistinguishable from training samples for each value of the condition.   

\begin{figure}[tbp]
    \centering
        \includegraphics[trim=20.6cm 0cm 10.4cm 0cm, clip,width=0.32\textwidth]{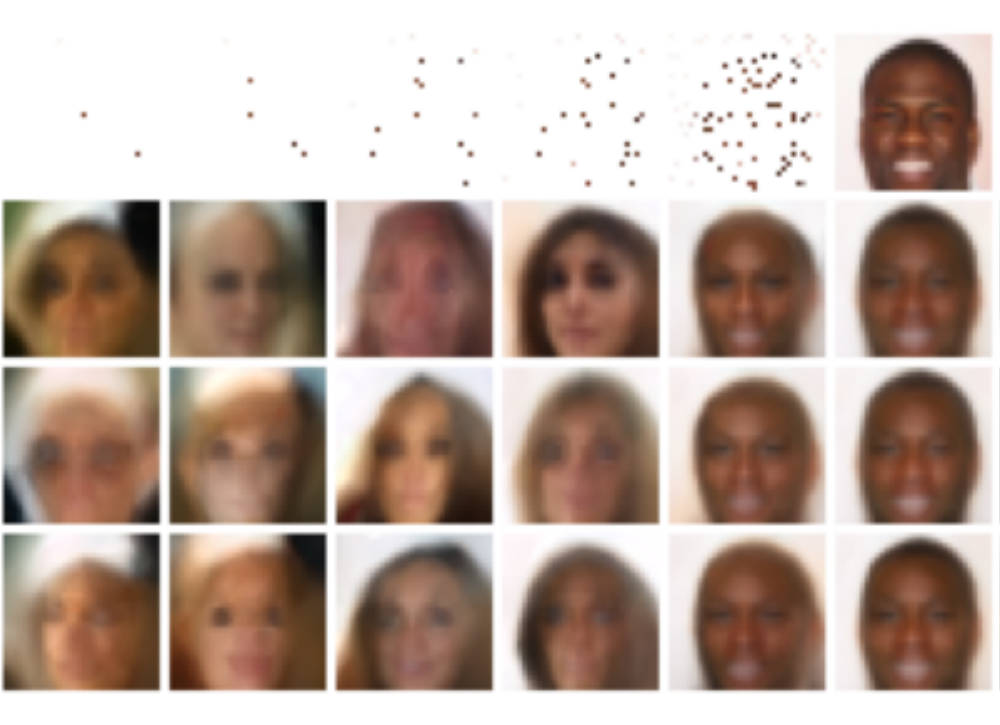}
        \includegraphics[trim=20.6cm 0cm 10.4cm 0cm, clip,width=0.32\textwidth]{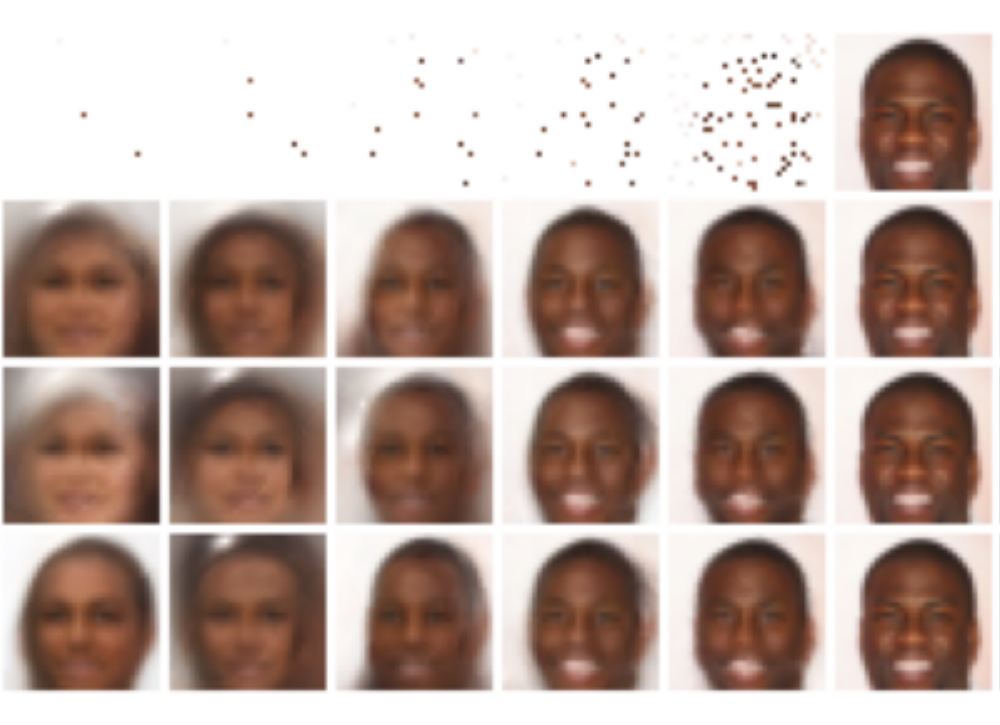}
        \includegraphics[trim=20.6cm 0cm 10.4cm 0cm, clip,width=0.32\textwidth]{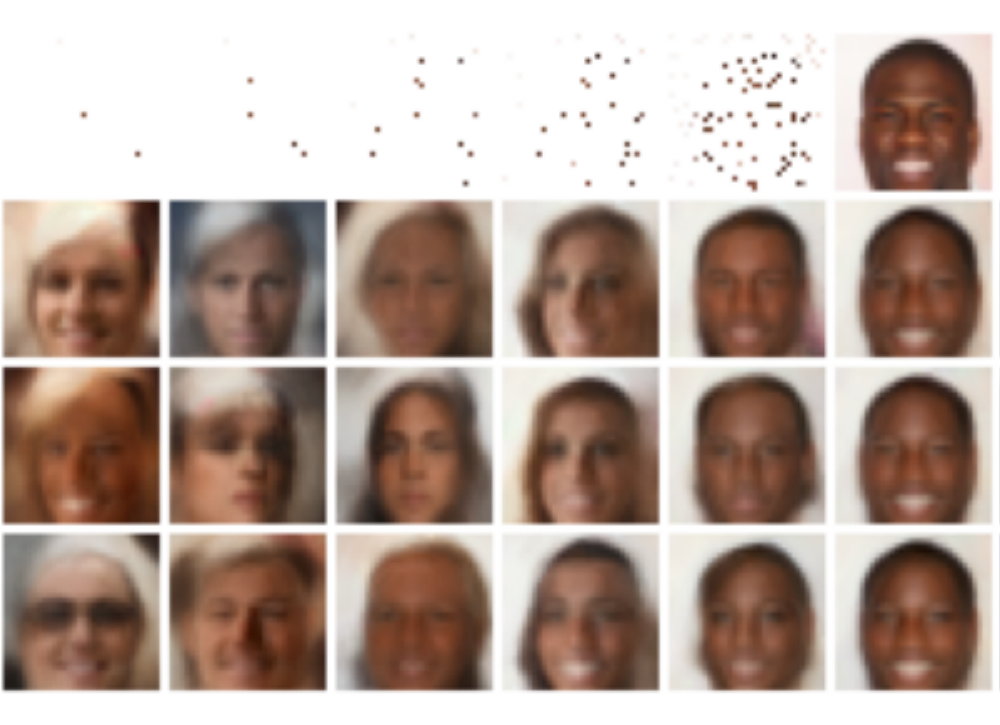}
        \includegraphics[trim=20.6cm 0cm 10.4cm 0cm, clip,width=0.32\textwidth]{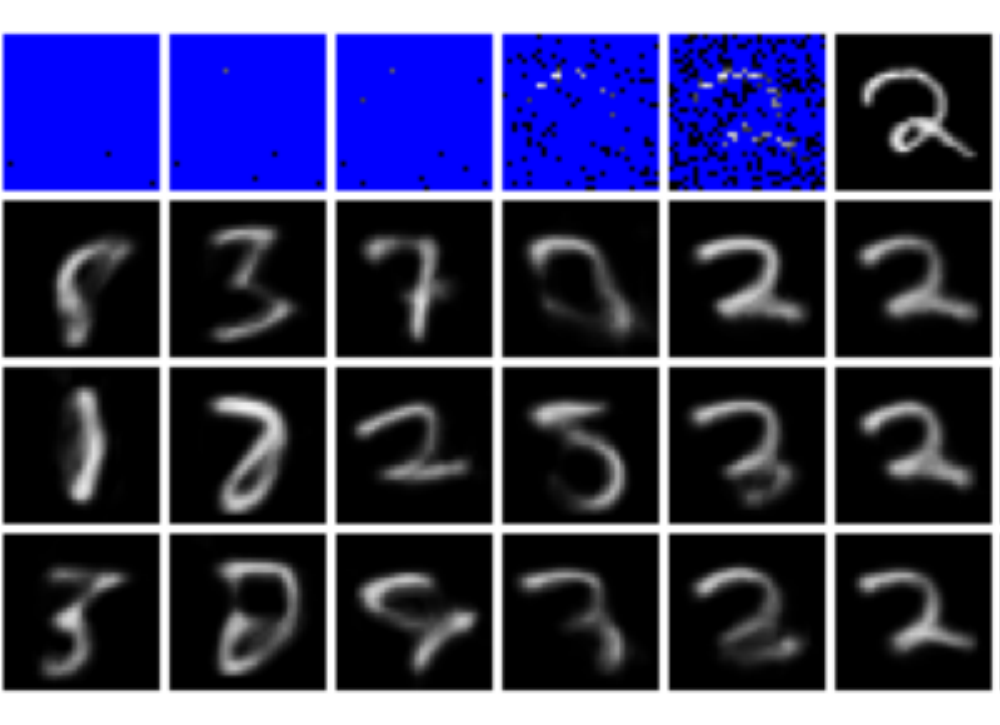}
        \includegraphics[trim=20.6cm 0cm 10.4cm 0cm, clip,width=0.32\textwidth]{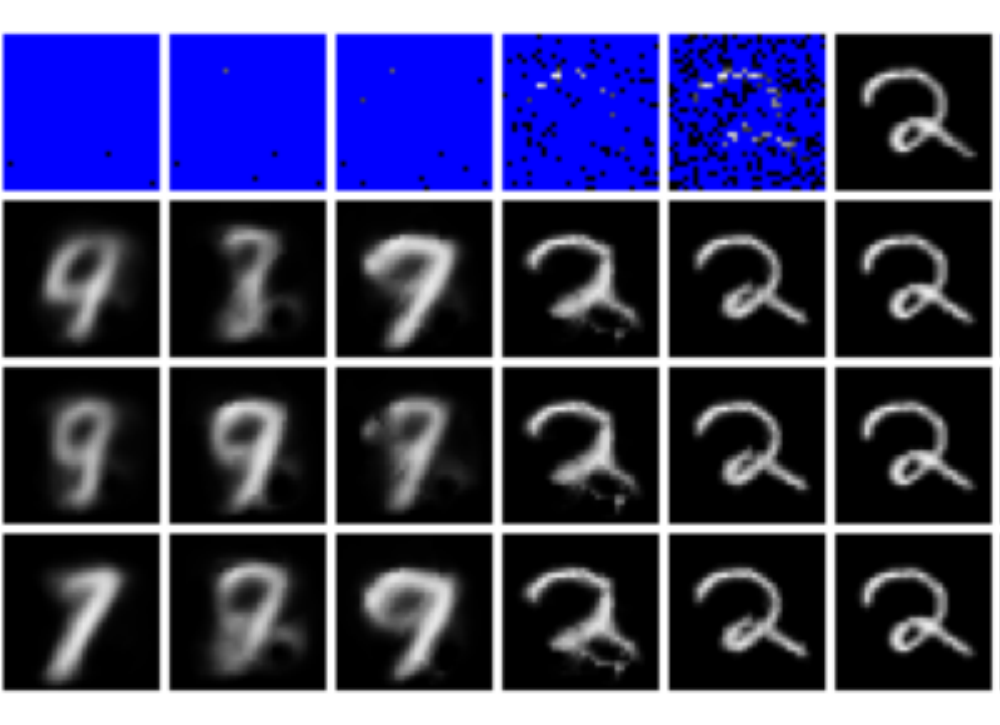}
        \includegraphics[trim=20.6cm 0cm 10.4cm 0cm, clip,width=0.32\textwidth]{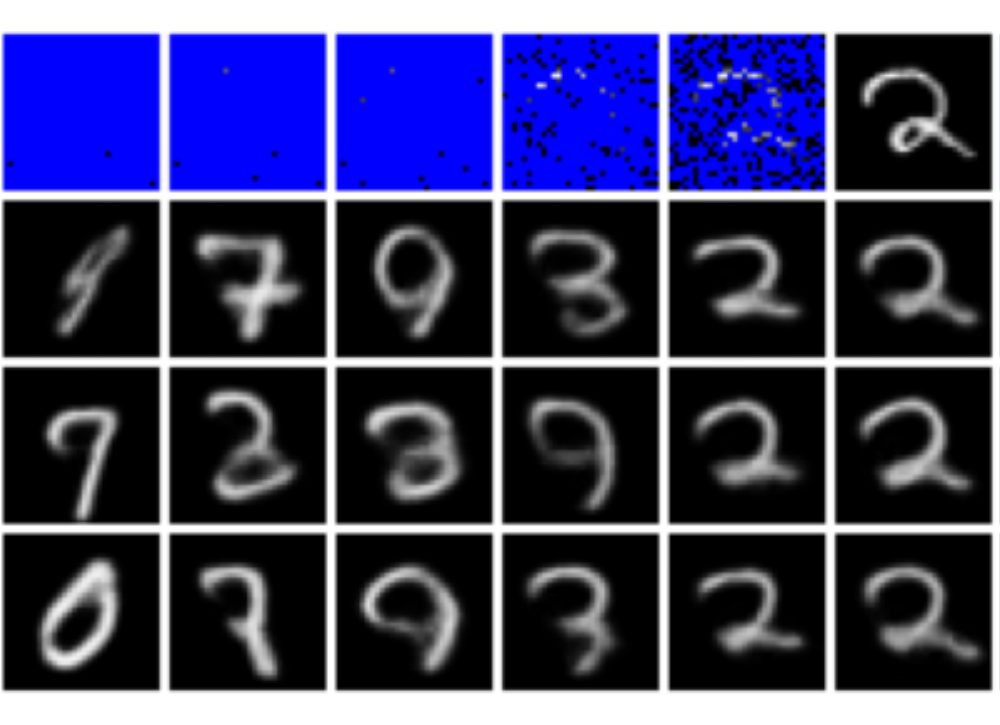}
 
    \caption{Representative image in-painting results for CelebA and MNIST.  From left to right, neural  process (NP) \citep{garnelo2018neural}, attentive neural process (ANP) \citep{kim2018attentive}, and ours.  Top rows show context sets of given pixels, ranging from very few pixels to all pixels.  In each panel the ground truth image (all pixels) is in the upper right corner.  The rows correspond to i.i.d.\ samples from the corresponding image completion model given only the pixels shown in the top row of the same column.  Our neural process with semi-implicit variational inference and max pooling produces results with the following characteristics: 1) the images generated with a small amount of contextual information are ``sharper'' and more face- and digit-like than NP results and 2) there is greater sample diversity across the i.i.d. samples than those from the ANP.  This kind of ``calibrated uncertainty'' is what we target throughout.}
    \label{fig:banner}
    \vspace{-0.2cm}
\end{figure}

Consider an amortized image in-painting task in which the objective is to fill in missing pixel values given a subset of observed pixel values. 
If the number and location of observed pixels is fixed, then a good conditional generative model should produce sharp-looking sample images, all of which should be compatible with the observed pixel values.  If the number and location of observed pixels is allowed to vary, the same should remain true for each set of observed pixels. 
Recent work on this problem has focused on reconstructing an entire image from as small a conditioning set as possible. As shown in \cref{fig:banner}, state-of-the-art methods  \citep{kim2018attentive} achieve high-quality reconstruction from as few as 30 conditioning pixels in a 1024-pixel image.

Our work starts by questioning whether reconstructing an image from a small subset of pixels is always the right objective. To illustrate, consider the image completion task on handwritten digits. A small set of pixels might, depending on their locations, rule out the possibility that the full image is, say, 1, 5, or 6. Human-like performance in this case would generate sharp-looking sample images for \emph{all} digits that are consistent with the observed pixels (i.e., 0, 2-4, and 7-9). Observing additional pixels will rule out successively more digits until the only remaining uncertainty pertains to stylistic details. The bottom-right panel of \cref{fig:banner} demonstrates this type of ``calibrated'' uncertainty. 

We argue that in addition to high-quality reconstruction based on large conditioning sets, amortized conditional inference methods should aim for {meaningful, calibrated uncertainty}, particularly for small conditioning sets. For different problems, this may mean different things. In this work, we focus on the image in-painting problem, and define well calibrated uncertainty to be a combination of two qualities: high sample diversity for small conditioning sets; and sharp-looking, realistic images for any size of conditioning set. As the size of the conditioning set grows, we expect the sample diversity to decrease and the quality of the images to increase. We note that this emphasis is different from the current trend in the literature, which has focused primarily on making sharp and accurate image completions when the size of the conditioning context is large \citep{kim2018attentive}.

To better understand and make progress toward our aim, we employ posterior predictive inference in a conditional generative latent-variable model, with a particular focus on neural processes (NPs) \citep{garnelo2018conditional,garnelo2018neural}. We find that particular architecture choices can result in markedly different performance. In order to understand this, we investigate posterior uncertainty in NP models, and we use our findings to establish new best practices for NP amortized inference artifacts with well-calibrated uncertainty. 
In particular, we demonstrate improvements arising from a combination of max pooling, a mixture variational distribution, and a ``normal'' amortized variational inference objective. 

\section{Amortized Inference for Conditional Generative Models}
Our work builds on amortized inference \citep{gershman2014amortized, kingma2014auto}, probabilistic meta-learning \citep{Gordon:etal:2018:versa}, and conditional generative models in the form of neural processes \citep{garnelo2018neural,kim2018attentive}. This section provides background.

\begin{figure}[tbp]
    \centering
    \includegraphics[width=0.7\linewidth]{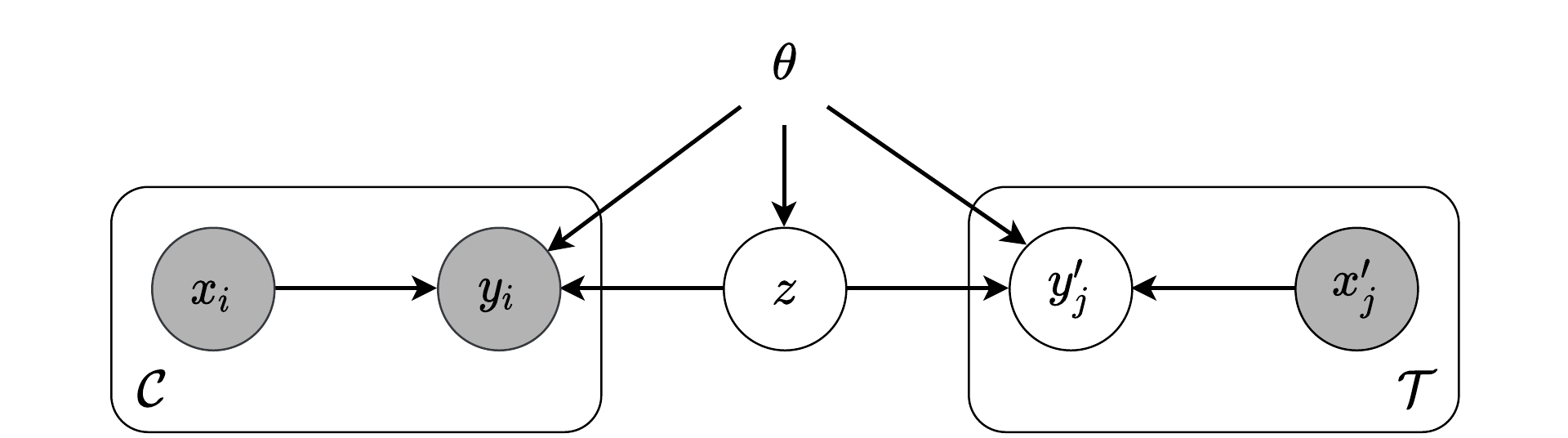}
    \caption{Graphical model for a {\em single} neural process task. $\gC$ is the task ``context'' set of input/output pairs $(x_i,y_i)$ and $\gT$ is a target set in which only the input values are known.} 
    \label{fig:ml_pip_graphical}
    \vspace{-0.2cm}
\end{figure}

Let $(\vx_{\gC}, \vy_{\gC}) = \{(x_i, y_i)\}_{i=1}^{n}$ and $(\vx_{\gT}, \vy_{\gT}) = \{(x'_j, y'_j)\}_{j=1}^{m}$ be a context set and target set respectively. In image in-painting, the context set input $\vx_{\gC}$ is a subset of an image’s pixel coordinates, the context set output $\vy_{\gC}$ are the corresponding pixel values (greyscale intensity or colors), the target set input $\vx_{\gT}$ is a set of pixel coordinates requiring in-painting, and the target set output $\vy_{\gT}$ is the corresponding set of target pixel values.  The corresponding graphical model is shown in  \cref{fig:ml_pip_graphical}.

The goal of amortized conditional inference is to rapidly approximate, at ``test time,'' the posterior predictive distribution
\begin{equation}
    p_\theta(\vy_\gT|\vx_\gT, \vx_\gC, \vy_\gC) = \int p_\theta(\vy_\gT|\vx_\gT, z) p_\theta(z|\vx_\gC, \vy_\gC) dz \;.\label{eq:post_predict}
\end{equation}
We can think of the latent variable $z$ %
as representing a problem-specific task-encoding.    The likelihood term $p_\theta(\vy_\gT|\vx_\gT, z)$ shows that the encoding parameterizes a regression model linking the target inputs to the target outputs. 
In the NP perspective, $z$ is a function and \cref{eq:post_predict} can be seen as integrating over the regression function itself, as in Gaussian process regression \citep{rasmussen2003gaussian}. 

\paragraph{Variational inference} There are two fundamental aims for amortized inference for conditional generative models: learning the model, parameterized by $\theta$, that produces good samples, and producing an amortization artifact, parameterized by $\phi$, that can be used to approximately solve \cref{eq:post_predict} quickly at test time. 
Variational inference techniques couple the two learning problems. 
Let $\vy$ and $\vx$ be task-specific output and input sets, respectively, and assume that at training time we know the values of $\vy$.  We can construct the usual single-training-task evidence lower bound (ELBO) as
\begin{equation}
    \log p_\theta(\vy|\vx) \geq {\textstyle\E_{\vz\sim q_\phi(z|\vx, \vy)} \left[\log \frac{p_\theta(\vy|z, \vx) p_\theta(z)}{q_\phi(z|\vx, \vy)} \right]} \;.
    \label{eq:np_elbo_intial}
\end{equation}
Summing over all training examples and optimizing \cref{eq:np_elbo_intial} with respect to $\phi$ learns an amortized inference artifact that takes a context set and returns a task embedding; optimizing with respect to $\theta$ learns a problem-specific generative model.  
Optimizing both simultaneously results in an amortized inference artifact bespoke to the overall problem domain.  

At test time the learned model and inference artifacts can be combined to perform amortized posterior predictive inference, approximating \cref{eq:post_predict} with
 \begin{equation}
    p_\theta(\vy_\gT|\vx_\gT, \vx_\gC, \vy_\gC) \approx\int p_\theta(\vy_\gT|\vx_\gT, z) q_\phi(z|\vx_\gC, \vy_\gC) dz \;. \label{eq:post_predict_approx}
\end{equation}
Crucially, given an input $(\vx_\gC,\vy_\gC)$, sampling from this distribution is as simple as sampling a task embedding $z$ from $q_\phi(z|\vx_\gC, \vy_\gC)$ and then passing the sampled $z$ to the generative model $p_\theta(\vy_\gT|\vx_\gT, z)$ to produce samples from the conditional generative model.

\paragraph{Meta-learning} The task-specific problem becomes a meta-learning problem when learning a regression model $\theta$ that performs well on \emph{multiple} tasks with the same graphical structure, trained on data for which the target outputs $\{y'_j\}$ are observed as well. 
In training our in-painting models, following conventions in the literature \citep{garnelo2018conditional,garnelo2018neural}, tasks are simply random-size subsets of random pixel locations $\vx$ and values $\vy$ from training set images.  %
This random subsetting of training images into context and target sets transforms this into a meta-learning problem, and  
the ``encoder'' $q_\phi(z|\vx, \vy)$ must learn to generalize over different context set sizes, with less posterior uncertainty as the context set size grows.  %

\paragraph{Neural processes} Our work builds on neural processes (NPs) \citep{garnelo2018conditional, garnelo2018neural}.  NPs are deep neural network  conditional generative models.  Multiple variants of NPs have been proposed \citep{garnelo2018conditional, garnelo2018neural, kim2018attentive}, and careful empirical comparisons between them  appear in the literature \citep{groverprobing,le2018empirical}. %

NPs employ an alternative training objective to \cref{eq:np_elbo_intial} arising from the fact that the graphical model in \cref{fig:ml_pip_graphical} allows a Bayesian update on the distribution of $z$, conditioning on the entire context set to produce a posterior $p_\theta(z|\vx_\gC, \vy_\gC)$. If the generative model is in a tractable family that allows analytic updates of this kind, then the NP objective corresponds to maximizing
\begin{equation}
    \E_{\vz\sim q_\phi(z|\vx_\gT, \vy_\gT)} \left[ \log {\textstyle\frac{p_\theta(\vy_\gT|z, \vx_\gT) p_\theta(z|\vx_\gC, \vy_\gC)}{q_\phi(z|\vx_\gT, \vy_\gT)}} \right] 
    \approx
    \E_{\vz\sim q_\phi(z|\vx_\gT, \vy_\gT)} \left[ \log {\textstyle\frac{p_\theta(\vy_\gT|z, \vx_\gT) \textcolor{blue}{q_\phi(z|\vx_\gC, \vy_\gC)}}{q_\phi(z|\vx_\gT, \vy_\gT)}} \right] %
    \label{eq:np_elbo}
\end{equation}
where replacing $p_\theta(z|\vx_\gC, \vy_\gC)$ with its variational approximation is typically necessary because most deep neural generative models have a computationally inaccessible posterior.
This ``NP objective'' can be trained end-to-end, optimizing for both $\phi$ and $\theta$ simultaneously, where the split of training data into context and target sets must vary in terms of context set size.  The choice of optimizing \cref{eq:np_elbo} instead of \cref{eq:np_elbo_intial} is largely empirical  \citep{le2018empirical}.

\section{Network architecture}

\begin{figure}[tbp]
    \centering
    \begin{subfigure}[b]{0.55\linewidth}
        \centering
        \includegraphics[width=\linewidth]{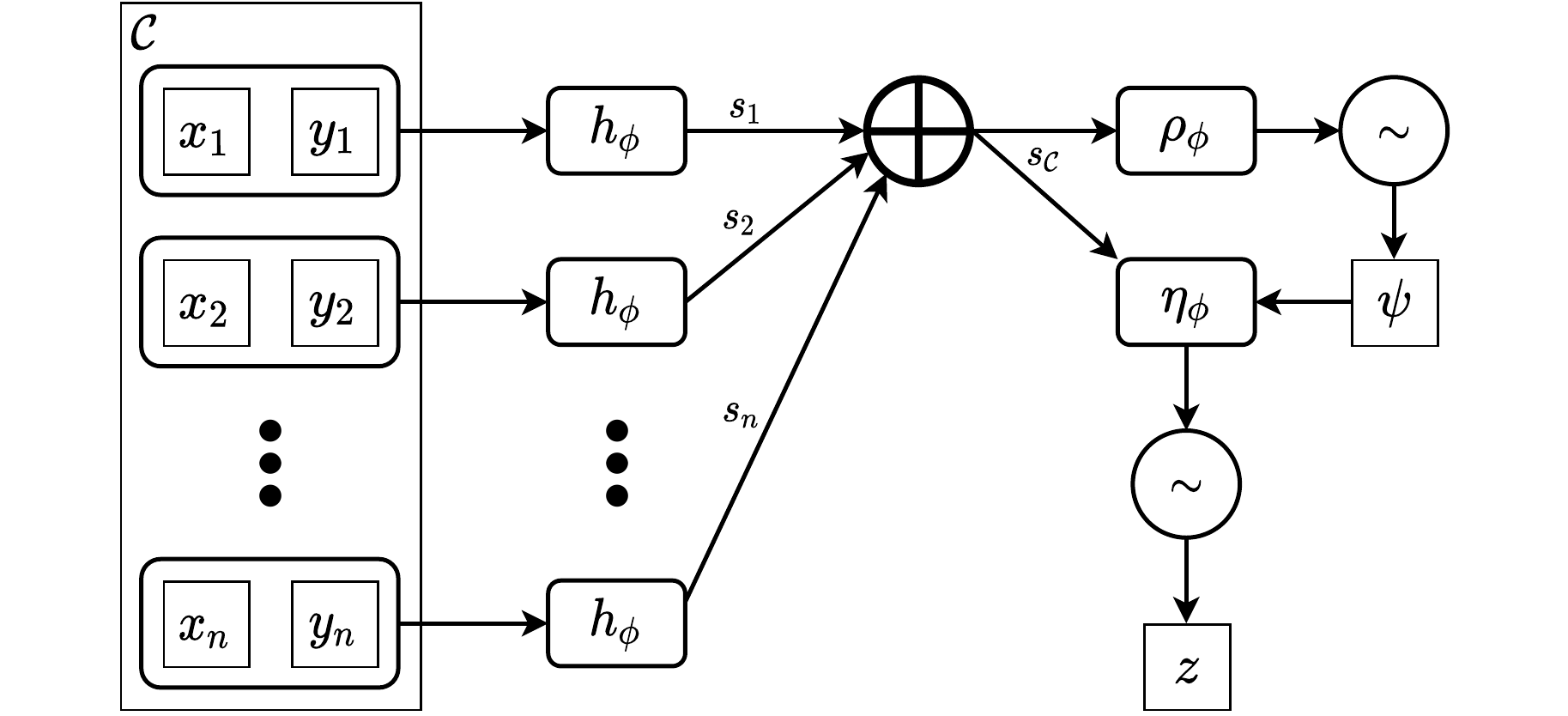}
        \caption{encoder}
        \label{fig:architecture:enc-hierarchical}
    \end{subfigure}
    \begin{subfigure}[b]{0.4\linewidth}
        \centering
        \includegraphics[width=\linewidth]{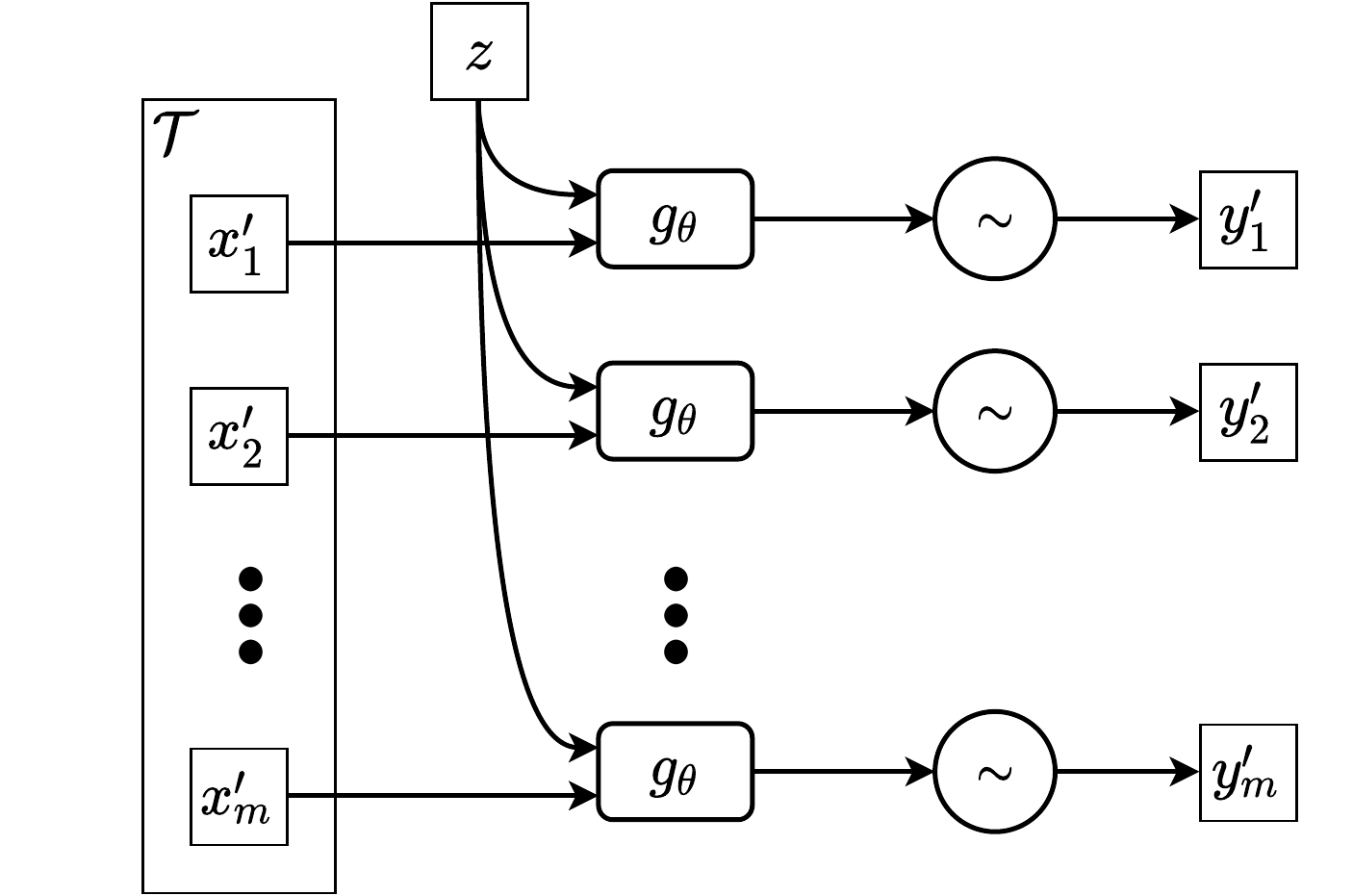}
        \caption{decoder}
        \label{fig:architecture:dec}
    \end{subfigure}
    \caption{Our modified neural process architecture.  The encoder  produces a permutation invariant embedding that parameterizes a stochastic task encoding $z$ as follows: features extracted from each element of the context set using neural net $h_\phi$ are pooled, then passed to other neural networks $\rho_\phi$ and $\eta_\phi$ that control the distribution over task embedding $z$.  The decoder uses such a task encoding along with embeddings of target inputs to produce the output distribution for each target input.}
    \label{fig:architecture}
\end{figure}

The network architectures we employ build on NPs, inspired by our findings from \cref{sec:uncertainty:discussion}. We describe them in detail in this section. 

\paragraph{Encoder}
The encoder $q_\phi(z|\vx_\gC, \vy_\gC)$ takes input observations from an i.i.d.\ model (see \cref{fig:ml_pip_graphical}, plate over $\gC$), and therefore its encoding of those observations must be permutation invariant if it is to be learned efficiently.  Our $q_\phi$, as in related NP work, has a  permutation-invariant architecture,
\begin{align*}
    s_i = h_\phi(x_i, y_i), \; 1 \leq i \leq n; \quad
    s_\gC = \oplus_{i=1}^{n} s_i; \quad 
    (\mu_\gC, \sigma_\gC) = \rho_\phi(s_\gC); \quad 
    q_\phi(z | \vx_\gC, \vy_\gC) = \gN(\mu_\gC, \sigma_{\gC}^{2}) \;. %
\end{align*}
Here $\rho_\phi$ and $h_\phi$ are neural networks and $\oplus$ is a permutation-invariant pooling operator.  \cref{fig:architecture} contains diagrams of a generalization of this encoder architecture (see below). The standard NP architecture uses mean pooling; motivated by our findings in \cref{sec:uncertainty:discussion}, we also employ max pooling. \textbf{}

\paragraph{Hierarchical Variational Inference}
In order to achieve better calibrated uncertainty in small context size regimes, a more flexible approximate posterior should be beneficial.  
Consider the MNIST experiment shown in \cref{fig:exp:combined}.  Intuitively, an encoder could learn to map from the context set to a one-dimensional discrete $z$ value that lends support only to  those digits that are compatible with the context pixel values at the given context pixel locations $(\vx_\gC, \vy_\gC)$.  This suggests that $q_\phi$ should be flexible enough to produce a multimodal distribution over $z$,  %
which can be encouraged by making $q_\phi$ a mixture and corresponds to a hierarchical variational distribution \citep{ranganath2016hierarchical,yin2018semi,sobolev2019importance}. 
Specifically, the encoder structure described above, augmented with a mixture variable is
\begin{equation}
    q_\phi(z| \vx, \vy) = \int q_\phi(\psi| \vx, \vy) q_\phi(z|\psi, \vx, \vy) d\psi \;.
    \label{eq:hierarchical-proposal}
\end{equation} 
This is shown in \cref{fig:architecture}. For parameter-learning, semi-implicit variational inference (SIVI) \citep{yin2018semi} constructs a tractable lower bound to the ELBO (See the Supplementary Material).
Our experimental findings suggest that the combination of max pooling and SIVI produce state-of-the-art high-quality and diverse samples from well calibrated posteriors, as illustrated in \cref{fig:exp:combined}.

\paragraph{Decoder} 
The deep neural network stochastic decoder in our work is standard and not a focus.  Like other NP work, the data generating conditional likelihood in our decoder is assumed to factorize in a conditionally independent way,
    $p_\theta(\vy_\gT|z, \vx_\gT) = \prod_{i=1}^m p_\theta(y'_i|z, x'_i),$
where $m$ is the size of the target set and $x'_i$ and $y'_i$ are a target set input and output respectively.  \cref{fig:architecture:dec} shows the decoder architecture, with the neural network $g_\theta$ the link function to a per pixel likelihood.

\section{Uncertainty in Neural Process models} \label{sec:uncertainty:discussion}
\begin{figure}[tbp]
    \centering
    \begin{subfigure}[b]{0.49\linewidth}
        \centering
        \includegraphics[width=\linewidth]{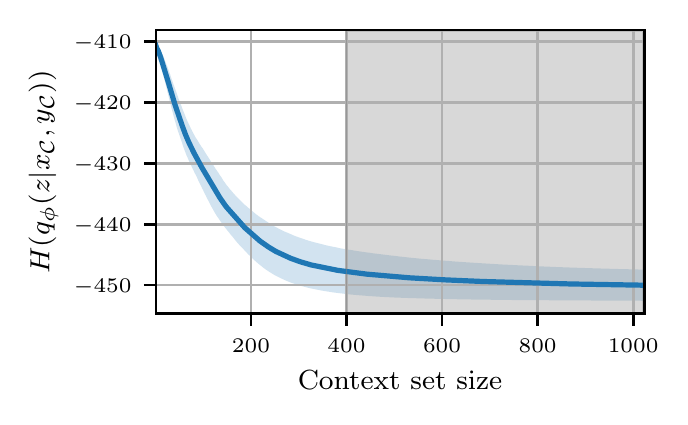}
        \caption{Variational posterior entropy}
        \label{fig:exp:max-posterior-contraction}
    \end{subfigure}
    \begin{subfigure}[b]{0.49\linewidth}
        \centering
        \includegraphics[width=\linewidth]{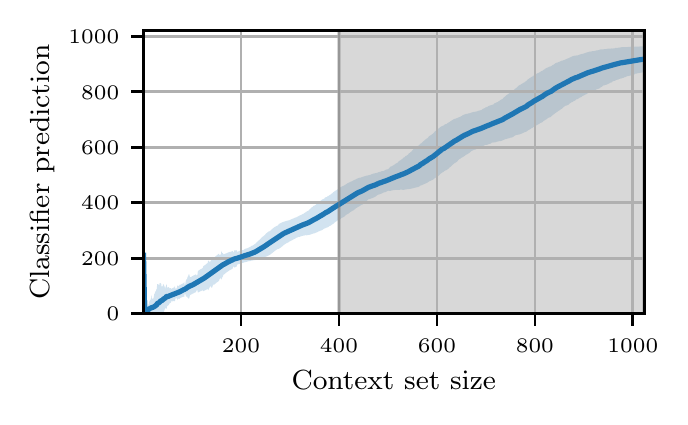}
        \caption{Classifier prediction}
        \label{fig:exp:max-n-classifier}
    \end{subfigure}
    \caption{Posterior contraction of $q_\phi(z|\vx_\gC, \vy_\gC)$ in a NP+max pooling model. (a) The entropy of $q_\phi(z|\vx_\gC, \vy_\gC)$ as a function of context set size, averaged over different tasks (images) and context sets. The gray shaded area in both plots indicates context set sizes that did not appear in the training data for the amortization artifact. (b) Predictions of a classifier trained to infer the context set size given only $s_\mathcal{C}$, the pooled embedding of a context set.  %
    Equivalent results for the standard NP+mean pooling encoder and for ANP appear in the Supplementary Material.}
    \label{fig:exp:max-n}
\end{figure}

In this section, we investigate how NP models handle uncertainty. 
A striking property of NP models is that as the size of the (random) context set increases, there is less sampling variation in target samples generated by passing $z\sim q_\phi(z | \vx_\gC, \vy_\gC)$ through the decoder. The samples shown in \cref{fig:banner} are the likelihood mean (hence a deterministic function of $z$), and so the reduced sampling variation can only be produced by decreased posterior uncertainty. Our experiments confirm this, as shown in \cref{fig:exp:max-posterior-contraction}: posterior uncertainty (as measured by entropy) decreases for increasing context size, \emph{even beyond the maximum training context size}. Such posterior contraction is a well-studied property of classical Bayesian inference and is a consequence of the inductive bias of exchangeable models. However, NP models do not have the same inductive bias explicitly built in. How do trained NP models exhibit posterior contraction without being explicitly designed to do so? How do they learn to do so during training?

\begin{figure}[tbp]
    \centering
    \begin{tabular}{ m{0.45\textwidth} m{0.52\textwidth} }
      \includegraphics[width=0.45\textwidth]{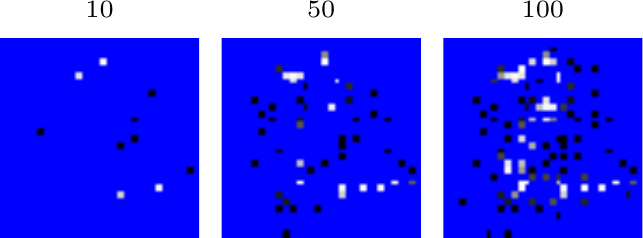} 
      &
      \includegraphics[width=0.51\textwidth]{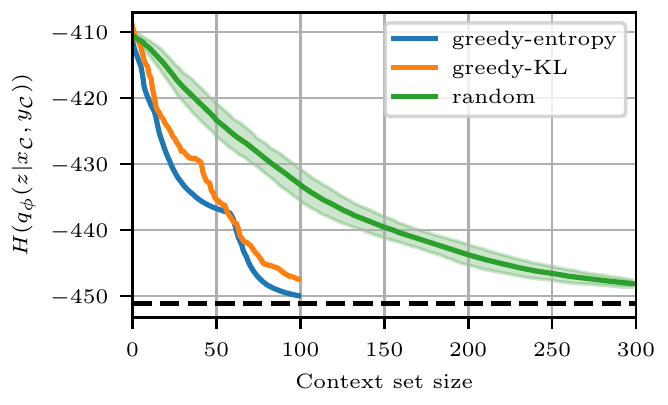} %
    \end{tabular}
    \caption{(Left) The first $\{10,50,100\}$ pixels greedily chosen to minimize $\KL(q_{\phi}(z|\vx,\vy) || q_\phi(z| \vx_\gC, \vy_\gC))$. These pixels are highly informative about $z$, but only a subset of them will appear the vast majority of random context sets. (Right) Posterior entropy decreasing as context size increases, for different methods of generating a context set: green is the average over 100 random context sets of each size; blue greedily chooses context pixels to minimize posterior entropy; and orange greedily minimizes $\KL(q_{\phi}(z|\vx,\vy) || q_\phi(z| \vx_\gC, \vy_\gC))$. The black dashed line represents the posterior entropy when conditioned on the full image.}
    \label{fig:uncert:entropyvssize}
\end{figure}

A simple hypothesis is that the network somehow transfers the context size through the pooling operation and into $\rho_\phi(s_\gC)$, which uses that information to set the posterior uncertainty. That hypothesis is supported by  \cref{fig:exp:max-n-classifier}, which shows the results of training a classifier to infer the context size given only $s_\gC$. However, consider that  within a randomly generated context set, some observations are more informative than others. %
For example, \cref{fig:uncert:entropyvssize} shows the first $\{10,50,100\}$ pixels of an MNIST 2, greedily chosen to minimize $\KL(q_{\phi}(z|\vx,\vy) || q_\phi(z| \vx_\gC, \vy_\gC))$. If $z$ is interpreted to represent, amongst other things, which digit the image contains, then a small subset of pixels determine which digits are possible. %

It is these highly informative pixels that drive posterior contraction in a trained NP. 
In a random context set, the number of highly informative pixels is random. For example, a max-pooled embedding saturates with the  $M$ most highly informative context pixels, where $M \leq d$, the dimension of embedding space. On average, a random context set of size $n$, taken from an image with $N$ pixels, will contain only  $\nicefrac{nM}{N}$ of the informative pixels.
In truth, \cref{fig:exp:max-n} displays how the information content of a context depends, on average, on the size of that context. Indeed, greedily choosing context pixels %
results in much faster contraction (\cref{fig:uncert:entropyvssize}).

\paragraph{Learning to contract} %
Posterior contraction is implicitly encouraged by the NP objective \cref{eq:np_elbo}. It can be rewritten as
\begin{align}
  \E_{\vz\sim q_\phi(z|\vx_\gT, \vy_\gT)} \left[ \log p_\theta(\vy_\gT|z, \vx_\gT) \right] - \KL( q_\phi(z|\vx_\gT, \vy_\gT) || q_\phi(z|\vx_\gC, \vy_\gC) ) 
  \label{eq:np_elbo_re} \;.
\end{align}
The first term encourages perfect reconstruction of $y_\gT$, and discourages large variations in $z\sim q_\phi(z|\vx_\gT, \vy_\gT)$, which would result in large variations in predictive log-likelihood. This effect is stronger for larger target sets since there are more target pixels to predict. %
In practice, $\gC \subset \gT$, so the first term also (indirectly) encourages posterior contraction for increasing context sizes. The second term, $\KL( q_\phi(z|\vx_\gT, \vy_\gT) || q_\phi(z|\vx_\gC, \vy_\gC) )$, reinforces the contraction by encouraging the context posterior to be close to the target posterior. 

Although the objective encourages posterior contraction, the network mechanisms for achieving contraction are not immediately clear. Ultimately, the details depend on interplay between the pixel embedding function, $h_\phi$, the pooling operation $\oplus$, and $\rho_\phi$. We focus on mean  and max pooling.

\paragraph{Max pooling} As the size of the context set increases, the max-pooled embedding $s_{\gC} = \oplus_{i=1}^n s_i$ is non-decreasing in $n$; in a trained NP model, $||s_{\gC}||$ will increase each time an informative pixel is added to the context set; %
it will continue increasing until the context embedding saturates at the full image embedding. At a high level, this property of max-pooling means that the $\sigma_\gC$ component of $\rho_\phi(s_\gC)$ has a relatively simple task: represent a function such that the posterior entropy is a decreasing function of all dimensions of the embedding space. An empirical demonstration that $\rho_\phi$ achieves this can be found in the Supplementary Material.

\paragraph{Mean pooling} For a fixed image, as the size of a random context set increases, its mean-pooled embedding will, on average, become closer to the full image embedding. Moreover, the mean-pooled embeddings of all possible context sets of the image are contained in the convex set whose hull is formed by (a subset of) the individual pixel embeddings. The $\sigma_\gC$ component of $\rho_\phi(s_\gC)$, then, must approximate a function such that the posterior entropy is a convex function on the convex set formed by individual pixel embeddings, with minimum at or near the full image embedding. Learning such a function across the embeddings of many training images seems a much harder learning task than that required by max pooling, which may explain the better performance of max pooling relative to mean pooling in NPs (see \cref{sec:experiments}).

\paragraph{Generalizing posterior contraction} 
Remarkably, trained NP-based models generalize their posterior contraction to context and target sizes not seen during training (see \cref{fig:exp:max-n}).  The discussion of posterior contraction in NPs using mean and max pooling in the previous paragraphs highlights a shared property: for both models, the pooled embeddings of all possible context sets that can be obtained from an image are in a convex set that is determined by a subset of possible context set embeddings. For max-pooling, the convex set is formed by the max-pooled embedding of the $M$ ``activation'' pixels. For mean-pooling, the convex set is obtained from the convex hull of the individual pixel embeddings. Furthermore, the full image embedding in both cases is contained in the convex set.
We conjecture that a sufficient condition for an NP image completion model to yield posterior contraction that generalizes to context sets of unseen size is as follows: For any image, the pooled embedding of every possible context set (which includes the full image) lies in a convex subset of the embedding space.

\section{Experimental evaluation} \label{sec:experiments}
\begin{figure}[!tb]
    \centering
    \begin{subfigure}[b]{0.45\linewidth}
        \centering
        \includegraphics[width=\linewidth]{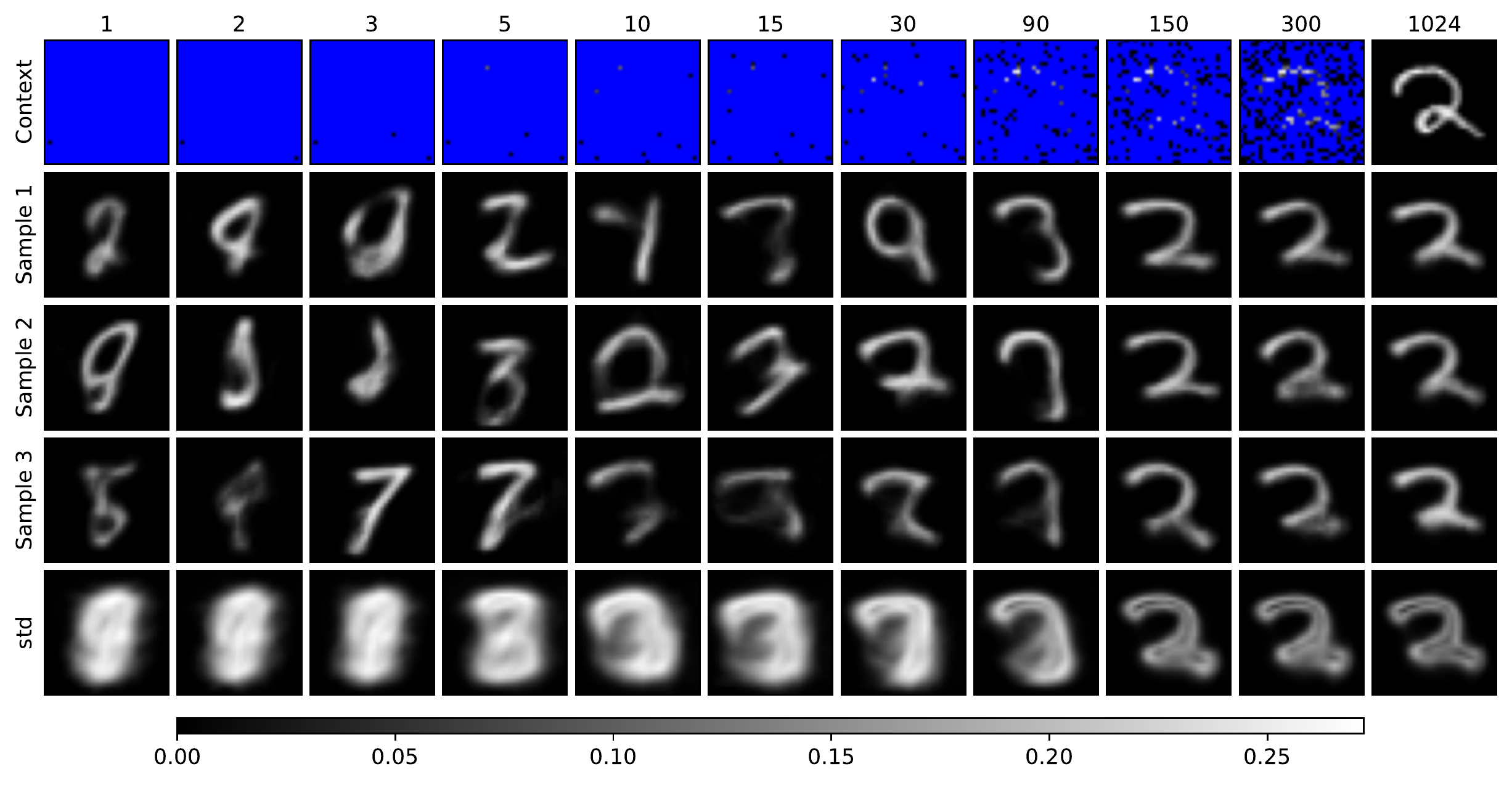}
        \caption{NP objective with average pooling}
        \label{fig:exp:mnist-np-avg}
    \end{subfigure}
        \begin{subfigure}[b]{0.45\linewidth}
        \centering
        \includegraphics[width=\linewidth]{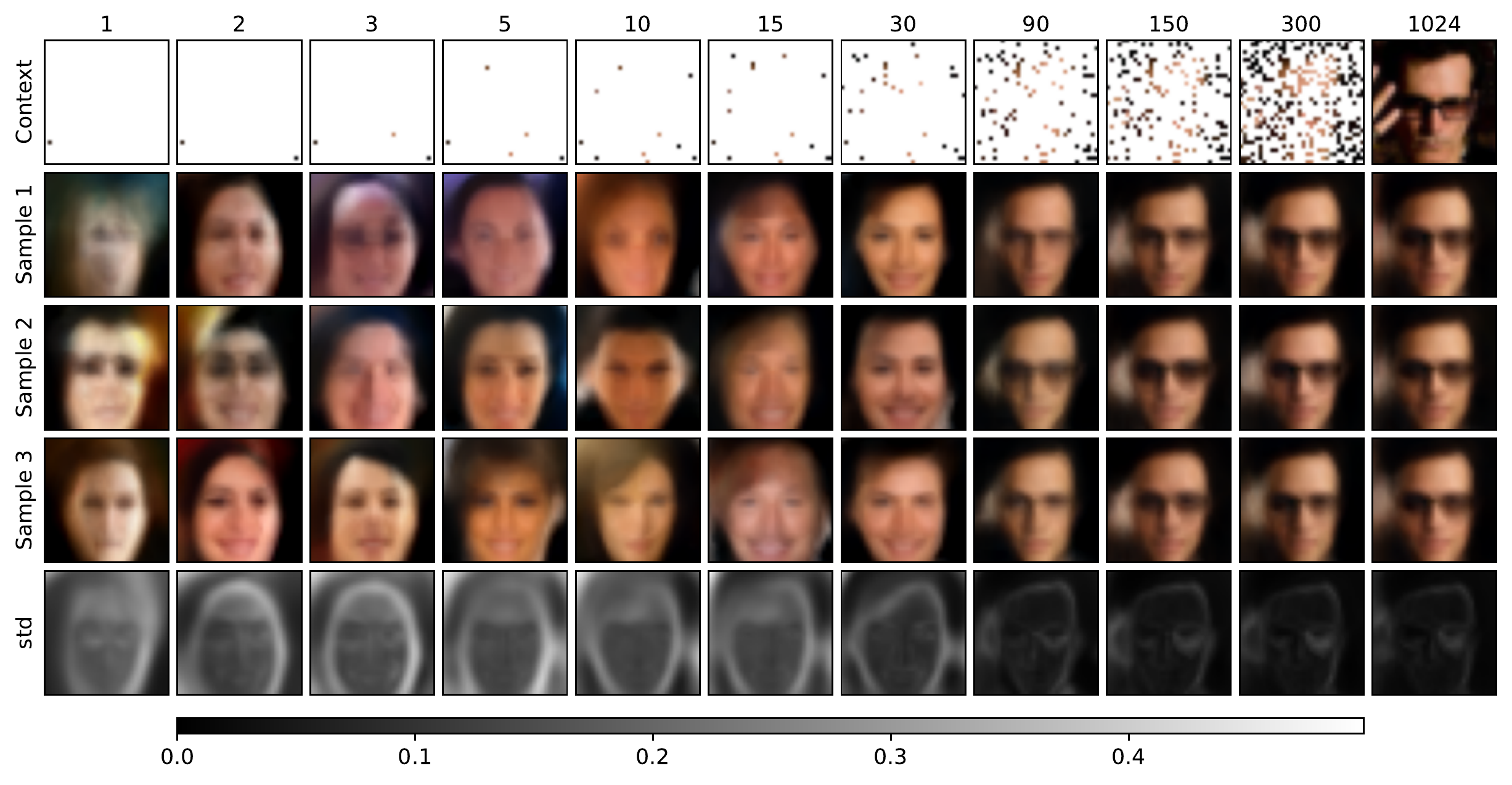}
        \caption{NP objective with average pooling}
        \label{fig:exp:celeba-np-avg}
    \end{subfigure}
    \hfill
        \begin{subfigure}[b]{0.45\linewidth}
        \centering
        \includegraphics[width=\linewidth]{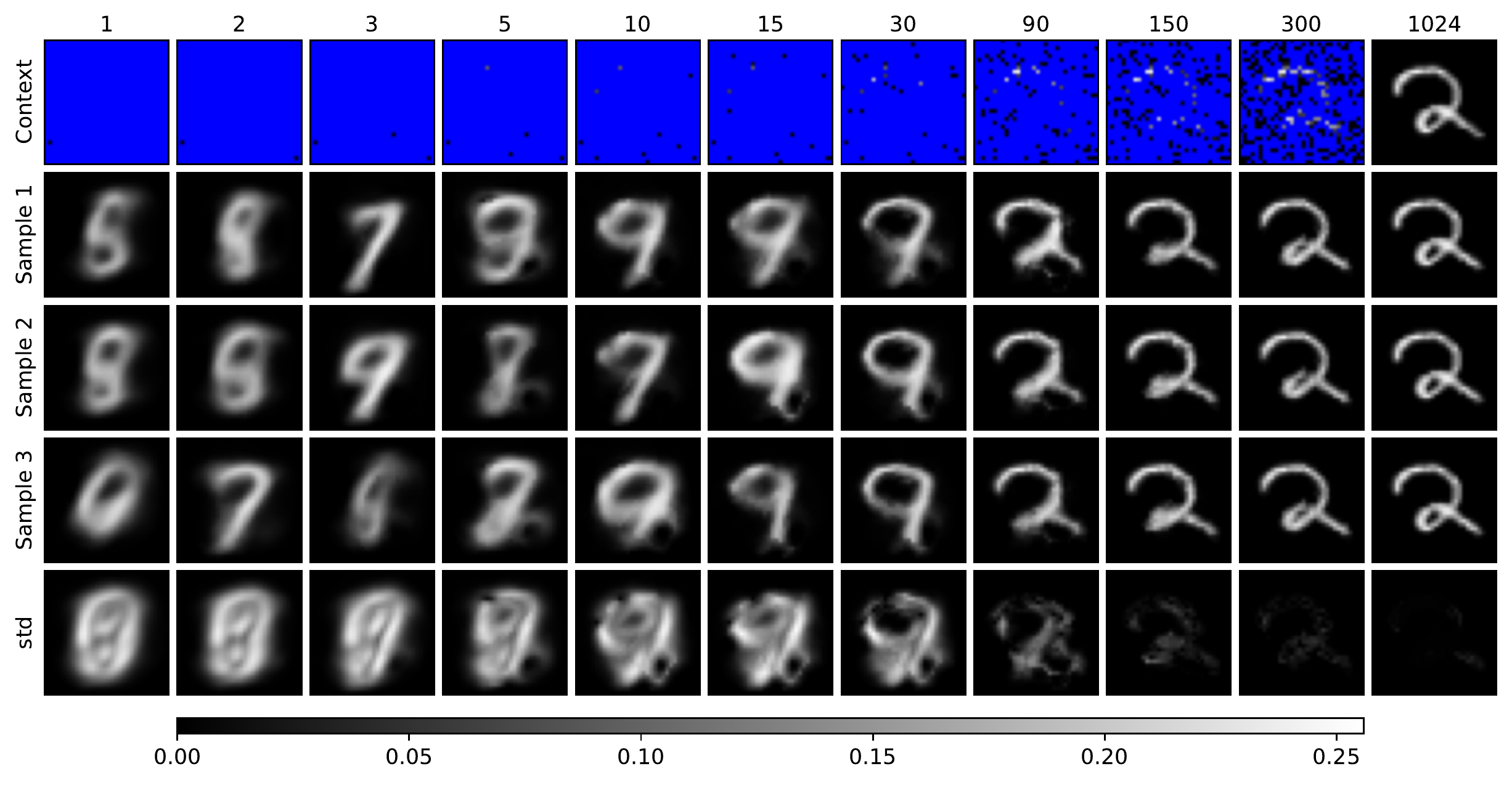}
        \caption{ANP}
        \label{fig:exp:mnist-anp-avg}
    \end{subfigure}
     \begin{subfigure}[b]{0.45\linewidth}
        \centering
        \includegraphics[width=\linewidth]{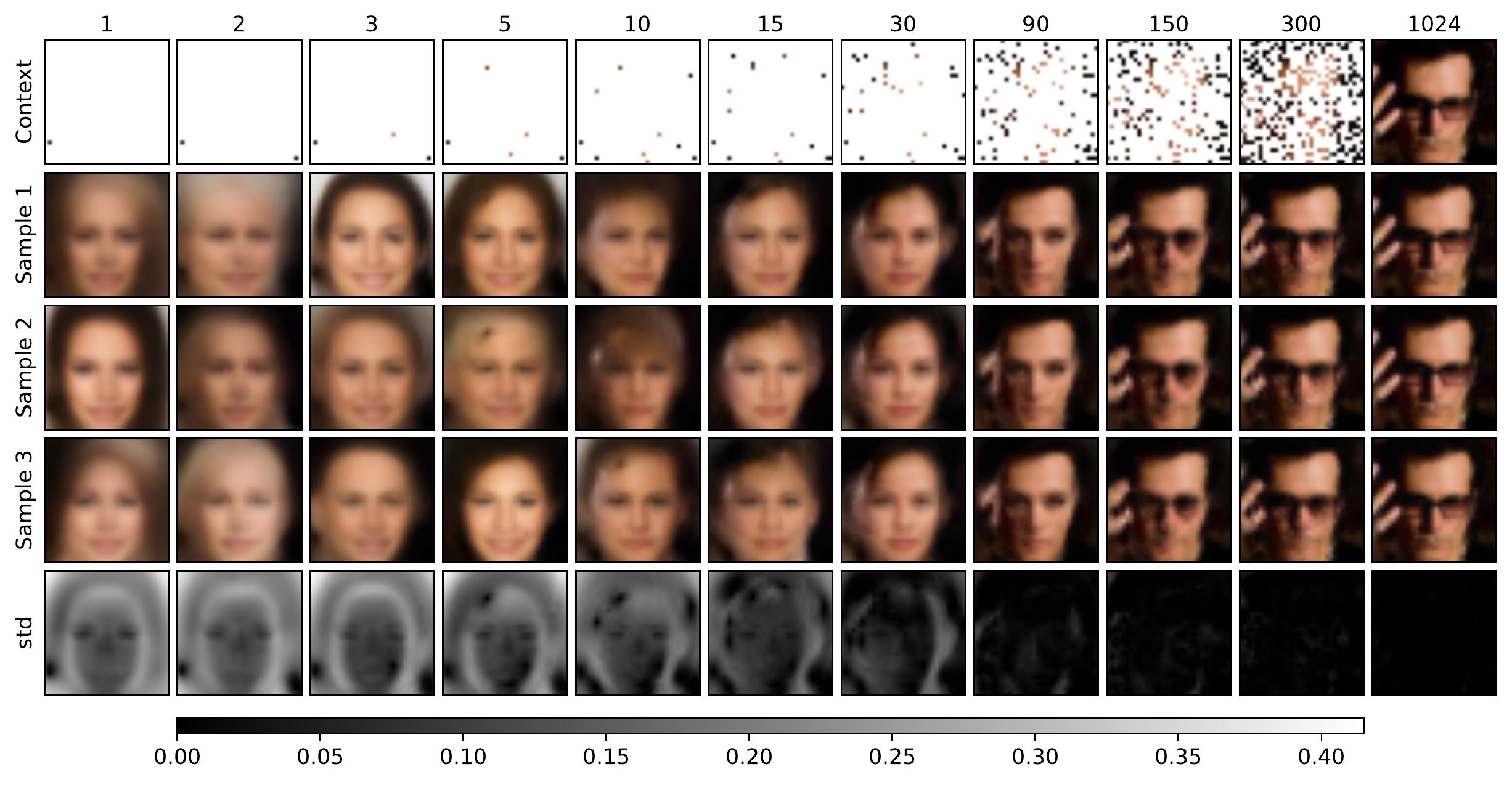}
        \caption{ANP}
        \label{fig:exp:celeba-anp-avg}
    \end{subfigure}
    \hfill
    \begin{subfigure}[b]{0.45\linewidth}
        \centering
        \includegraphics[width=\linewidth]{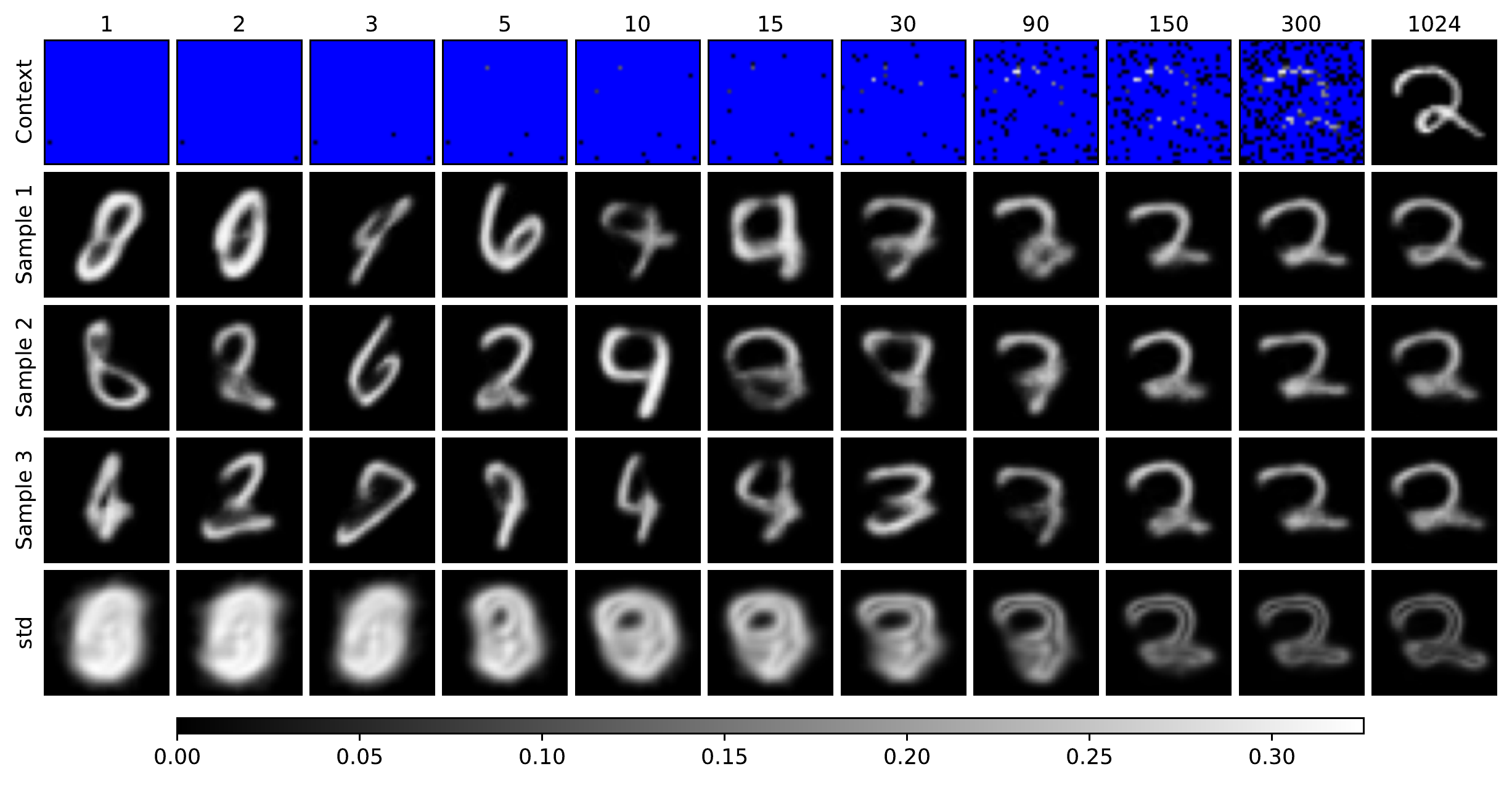}
        \caption{NP+SIVI objective with max pooling}
        \label{fig:exp:mnist-sivi-max}
    \end{subfigure}
    \begin{subfigure}[b]{0.45\linewidth}
        \centering
        \includegraphics[width=\linewidth]{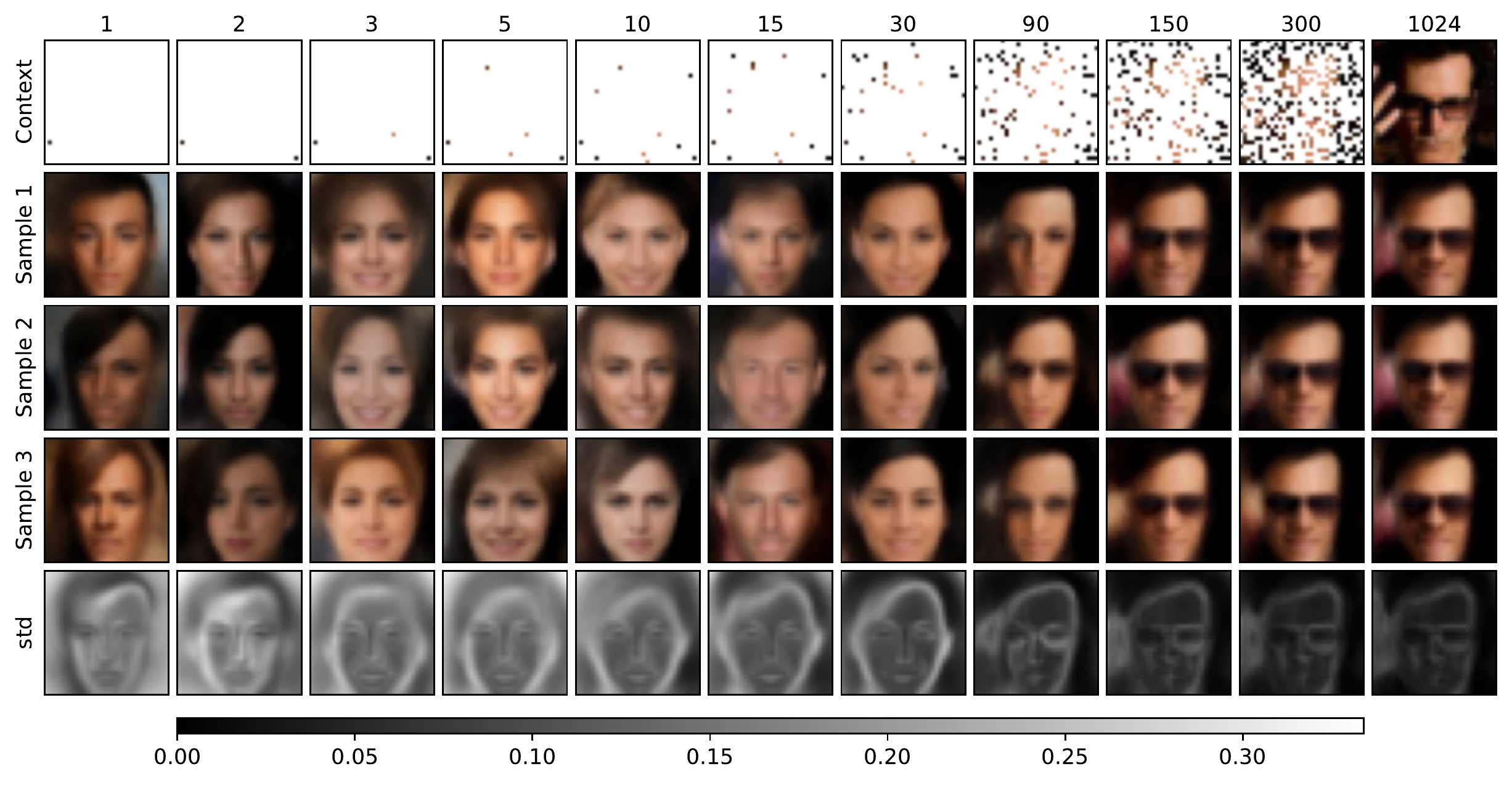}
        \caption{NP+SIVI objective with max pooling}
        \label{fig:exp:celeba-sivi-max}
    \end{subfigure}
    \hfill
    \caption{
    Example MNIST and CelebA image completion tasks, for each of three NP methods. %
    The following guide applies to each block. 
    The top row shows context sets of different sizes (context sets are exactly the same for all methods), i.e., one task per column. The ground truth image is in the upper right corner. The rows correspond to the mean function produced by $g_\theta$ for different sampled values of $z$. %
    The bottom row shows an empirical estimate of the standard deviation of the mean function from 1000 draws of $z$, a direct visualization of the uncertainty encoding. %
    }
    \label{fig:exp:combined}
\end{figure}

We follow the experimental setup of \citet{garnelo2018neural}, where images are interpreted as functions that map pixel locations to color values, and image in-painting is framed as an amortized predictive inference task where the latent image-specific regression function needs to be inferred from a small context set of provided pixel values and locations. %
For ease of comparison to prior work, we use the same MNIST~\citep{lecun1998gradient} and CelebA~\citep{liu2015deep} datasets.  %
Specific architecture details for all networks are provided in the Supplementary Materials and open-source code for all experiments will be released at the time of publication.

\paragraph{Qualitative Results} \label{sec:exp:qualitative}
\cref{fig:exp:combined} shows qualitative image in-painting results for MNIST and CelebA images.
It is apparent in both contexts that ANPs perform poorly when the context set is small, despite the superior sharpness of their reconstructions when given large context sets. The sets of digits and faces that ANPs produce are not sharp, realistic, nor diverse.
On the other hand, their predecessor, NP (with mean pooling), arguably exhibits more diversity but suffers at all context sizes in terms of realism of the images. 
Our NP+SIVI with max pooling approach produces results with two important characteristics: 1) the images generated with a small amount of contextual information are sharper and more realistic; and 2) there is high context-set-compatible variability across the i.i.d.\ samples. 
These qualitative results demonstrate that max pooling plus the SIVI objective result in posterior mean functions that are sharper and more appropriately diverse, except in the high context set size regime where diversity does not matter and ANP produces much sharper images.  Space limitations prohibit showing  large collections of samples where the qualitative differences are even more readily apparent.  The Supplementary Material contains more comprehensive examples.

\begin{table}[bp]
\begin{center}
\caption{Predictive held-out test log-likelihood}
\label{table:test_metrics}
\begin{tabular}{c|cc}
\multicolumn{1}{c}{\bf Method}  &{\bf MNIST} & {\bf CelebA}
\\ \hline\hline
NP+mean                      & $0.96 \pm 0.12$ & $2.91 \pm 0.30$\\
ANP+mean     & $0.55 \pm 0.12$ & $1.81 \pm 0.18$\\ %
\end{tabular}
\begin{tabular}{c|cc}
\multicolumn{1}{c}{\bf Method}  &{\bf MNIST} & {\bf CelebA}
\\ \hline\hline
NP+max                      & $1.07 \pm 0.11$ & $3.17 \pm 0.30$\\
SIVI+max                    & $0.99 \pm 0.25$ & $2.99 \pm 0.39$
\end{tabular}
\end{center}
\vspace{-0.6cm}
\end{table}

\paragraph{Quantitative Results} \label{sec:exp:quantitative}
Quantitatively assessing posterior predictive calibration is  an open problem \citep{salimans2016improved,heusel2017gans}.
\Cref{table:test_metrics}  reports, for the different architectures we consider, predictive held out test-data log-likelihoods averaged over 10,000 MNIST  and 19,962 CelebA test images respectively.  While the reported results make it clear that max pooling improves held-out test likelihood, %
likelihood alone does not provide a direct measure of sample quality nor diversity.  It simply measures how much mass is put on each ground-truth completion.  

\begin{figure}[btp]
    \centering
    \includegraphics[width=\linewidth]{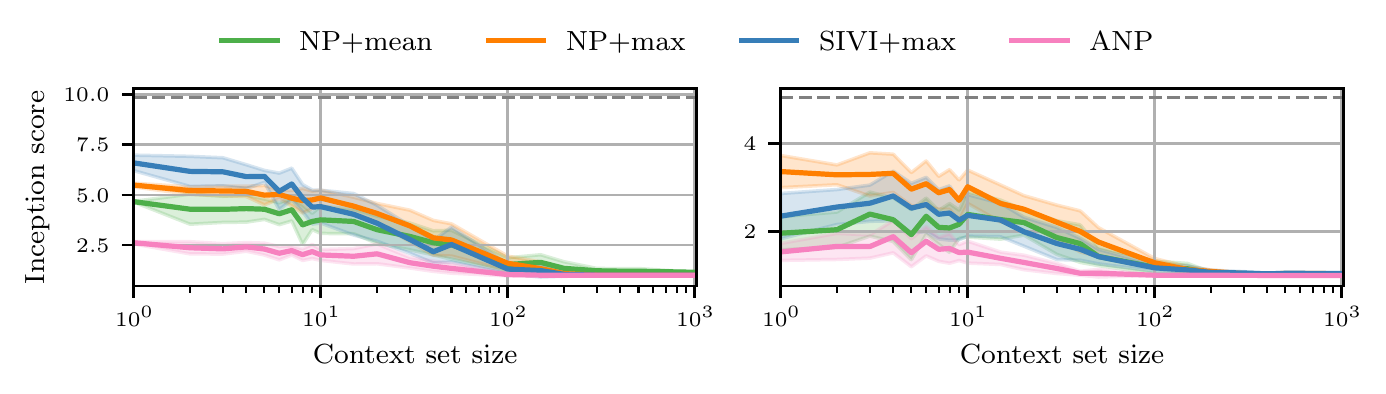}
    \caption{Inception scores of conditional samples (left, MNIST; right CelebA). 
    }
    \label{fig:exp:inception}
\end{figure}

Borrowing from the generative adversarial networks community, who have faced the similar  problems of how to quantitatively evaluate models via examination of the samples they generate, 
we compute inception scores \citep{salimans2016improved} using  conditionally generated samples for different context set sizes for all of the considered NP architectures and report them in \cref{fig:exp:inception}. However, since inception scores are based on classification outputs of inception network \citep{szegedy2016rethinking}, an ImageNet \citep{imagenet_cvpr09} classifier, it is known to give misleading results when applied to other image domains \citep{barratt2018note} including MNIST and CelebA. We therefore use trained MNIST and CelebA classifiers \citep{he2016deep} in place of inception network. (See Supplementary Materials for details.) 
The images used to create the results in \cref{fig:exp:inception} are the same as in \cref{fig:exp:combined} and the sequence of context sets considered include the ones in \cref{fig:exp:combined}. For each context set size, the reported inception scores are aggregated over 10 different randomly chosen context sets.  The dark gray dashed lines are the inception scores of training samples and represent the maximum one might hope to achieve at a context set size of zero (these plots start at one).

For small context sets, an optimally calibrated model should have high uncertainty and therefore generate samples with high diversity, resulting in high inception scores as observed. As the context set grows, sample diversity should be reduced, resulting in lower scores.  Here again, architectures using max pooling produce large gains in inception score in low-context size settings.  Whether the addition of SIVI is helpful is less clear here. Nonetheless, the inception score is again only correlated with the qualitative gains we observe in \cref{fig:exp:combined}.

\section{Conclusion}

The contributions we report in this paper include suggested neural process architectures (max pooling, no deterministic path) and objectives (regular amortized inference versus~the heuristic NP objective, SIVI versus~non-mixture variational family) that produce qualitatively better calibrated posteriors, particularly in low context cardinality settings.  %
We provide empirical evidence of how natural posterior contraction may be facilitated by the neural process architecture.  
Finally, we establish quantitative evidence that shows improvements in neural process posterior predictive performance and highlight the need for better metrics for quantitatively evaluating posterior calibration.

We remind the reader that this work, like most other deep learning work, highlights the impact of varying only a small subset of the dimensions of architecture and objective degrees of freedom.  We found that, for instance, simply making $\rho_\phi$ deeper than that reported in the literature improved baseline results substantially. The choice of learning rate also had a large impact on the relative gap between the reported alternatives.  We report what we believe to be the most robust configuration across all the configurations that we explored: max pooling and SIVI consistently improve performance.

\subsubsection*{Acknowledgments}
SN, KC and FW are supported by Support for Teams to Advance Interdisciplinary Research (STAIR) grant. SN and FW additionally acknowledge the support of the Natural Sciences and Engineering Research Council of Canada (NSERC), the Canada CIFAR AI Chairs Program, and the Intel Parallel Computing Centers program. This material is based upon work supported by the United States Air Force Research Laboratory (AFRL) under the Defense Advanced Research Projects Agency (DARPA) Data Driven Discovery Models (D3M) program (Contract No. FA8750-19-2-0222) and Learning with Less Labels (LwLL) program (Contract No.FA8750-19-C-0515). Additional support was provided by UBC's Composites Research Network (CRN), Data Science Institute (DSI) grants. BBR acknowledges the support of NSERC. This research was enabled in part by technical support and computational resources provided by WestGrid (https://www.westgrid.ca/) and Compute Canada (www.computecanada.ca).

\newpage

\bibliography{refs}
\bibliographystyle{iclr2021_conference}

\clearpage
\appendix
\section{Image inpainting results} \label{app:inpainting}
\begin{figure}[htbp]
    \centering
        \includegraphics[trim=20.6cm 0cm 10.4cm 0cm, clip,width=0.32\textwidth]{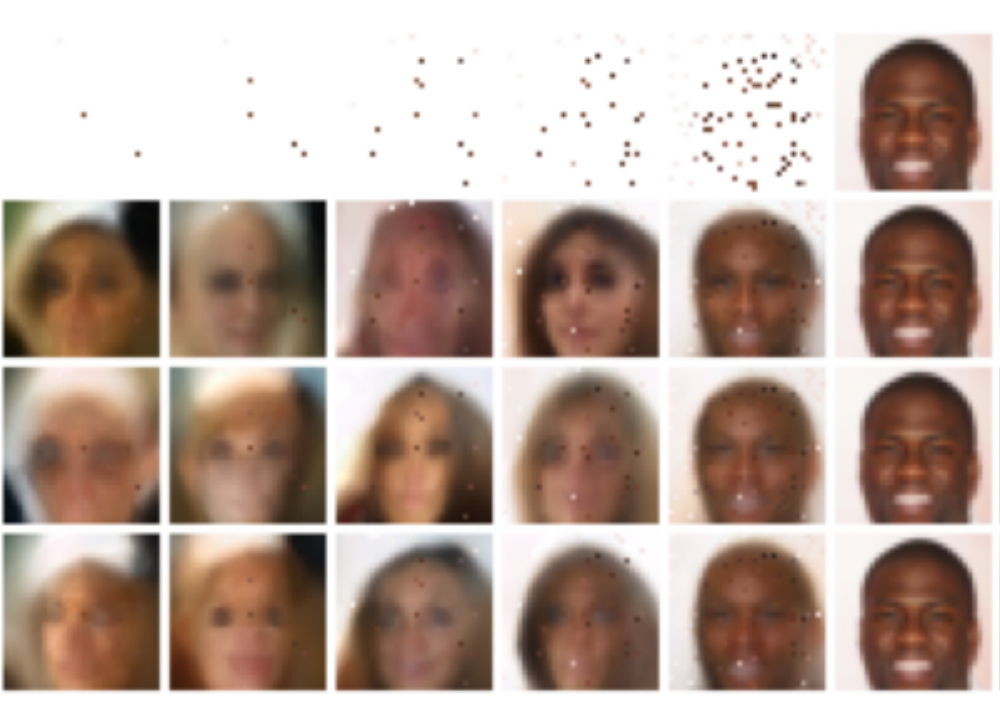}
        \includegraphics[trim=20.6cm 0cm 10.4cm 0cm, clip,width=0.32\textwidth]{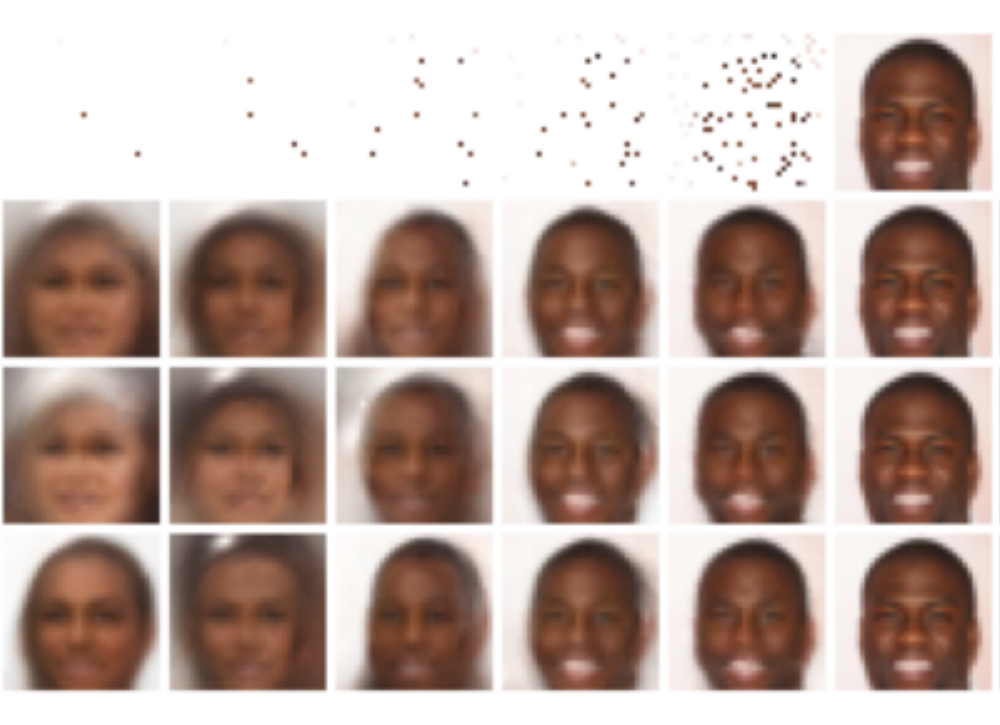}
         \includegraphics[trim=20.6cm 0cm 10.4cm 0cm, clip,width=0.32\textwidth]{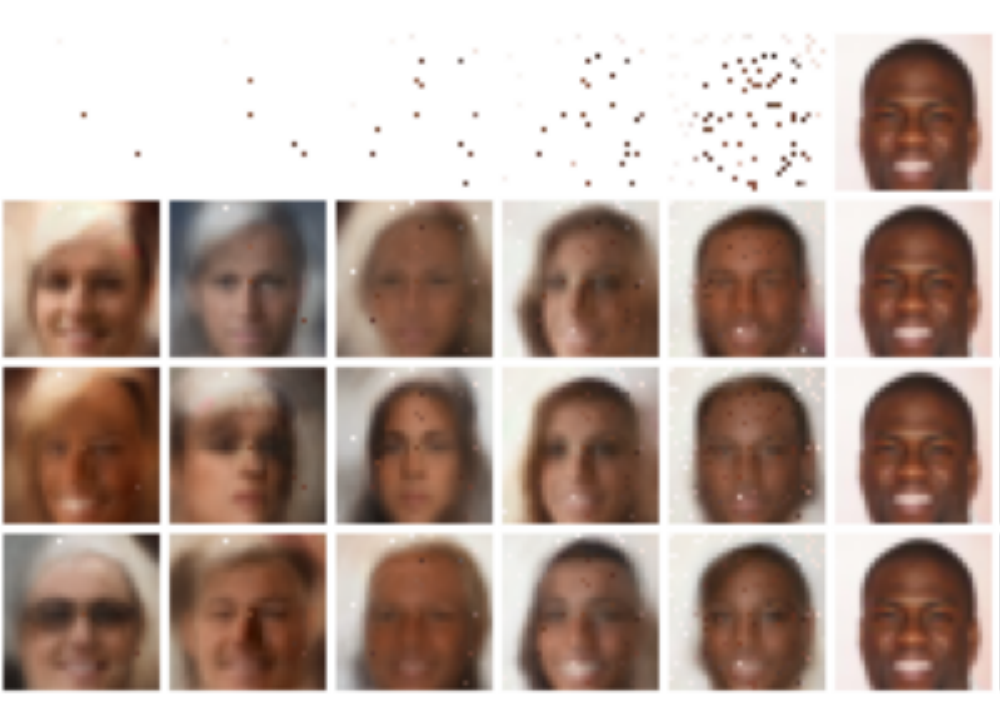}
        \includegraphics[trim=20.6cm 0cm 10.4cm 0cm, clip,width=0.32\textwidth]{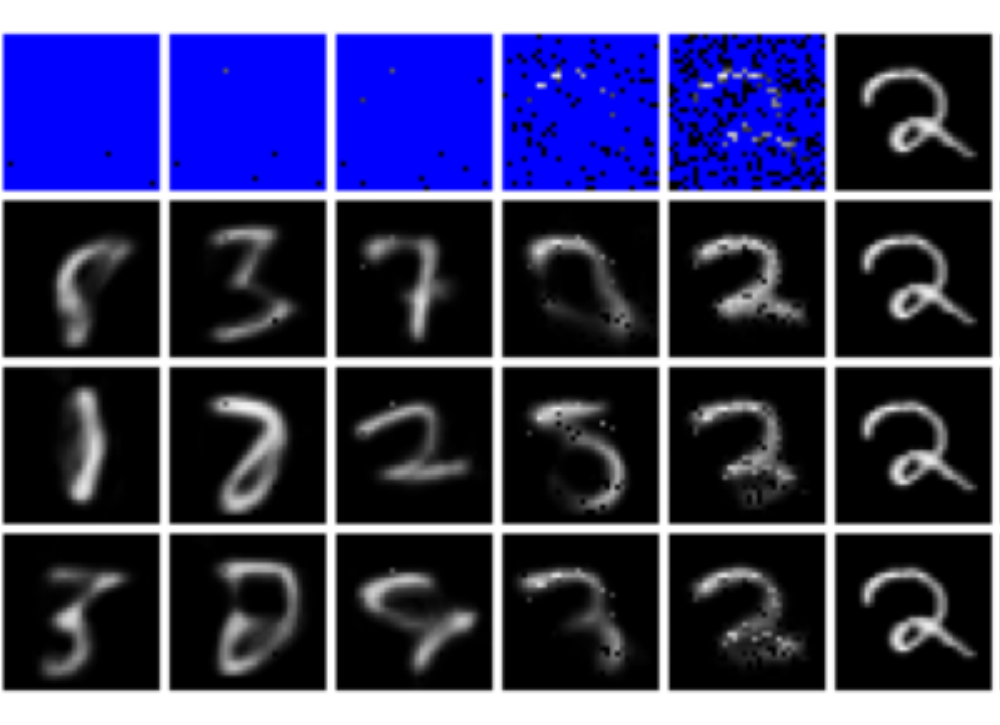}
        \includegraphics[trim=20.6cm 0cm 10.4cm 0cm, clip,width=0.32\textwidth]{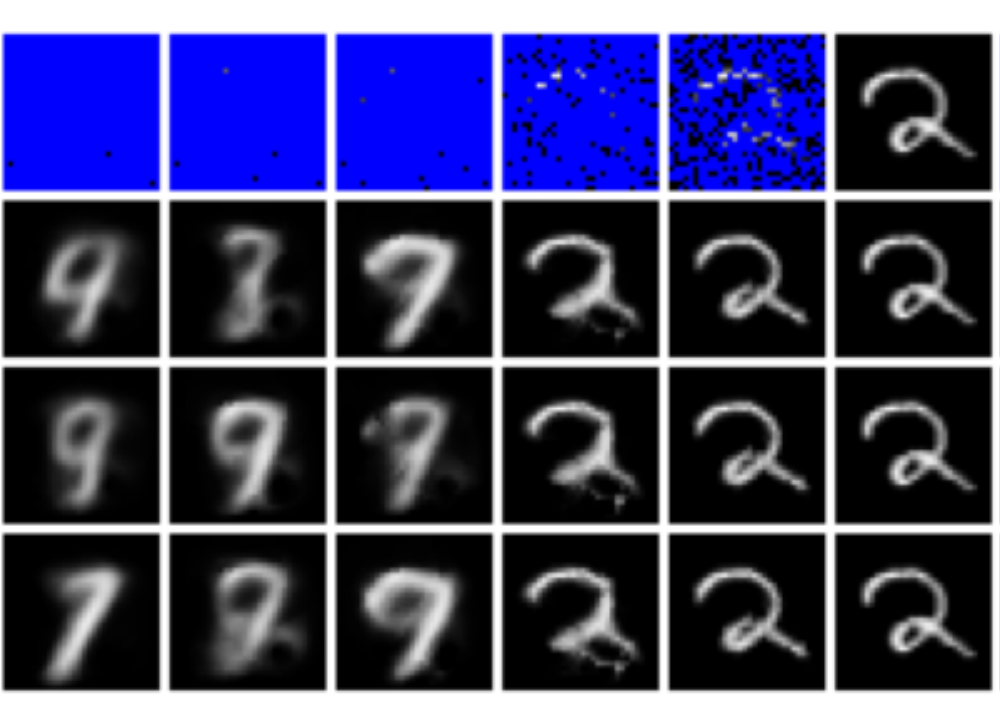}
        \includegraphics[trim=20.6cm 0cm 10.4cm 0cm, clip,width=0.32\textwidth]{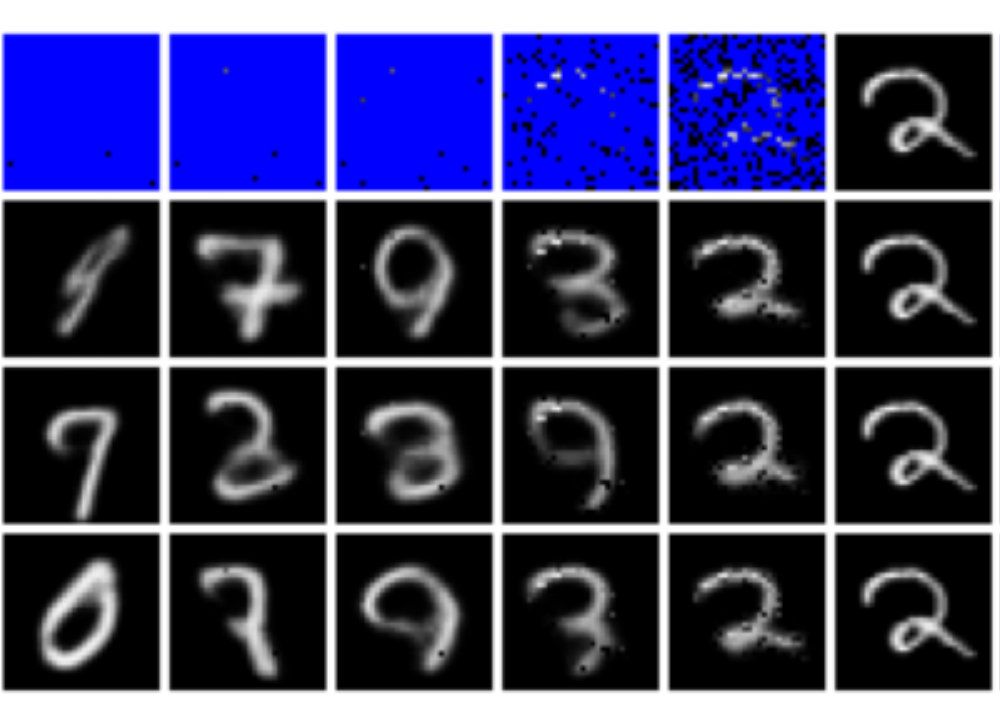}
    \caption{Image inpainting results when the context set is directly copied to the shown output and the model is queried for the rest of image pixels. From left to right, NP, ANP, and ours (NP+SIVI+max pooling). As non-attentive models are known to underfit on the context points, the final results for them are not smooth. ANP improves on this aspect.}
    \label{fig:inpainting-fixcontext}
\end{figure}

\cref{fig:inpainting-fixcontext} shows the inpainting results from different methods when the context set is carried to the output. In other words, inference is done via Equation~\ref{eq:post_predict} when the target and context sets are disjoint and the given $\vy_\gC$ is directly copied to the shown output, instead of asking the model to predict values of $\vy_\gC$.

\section{Architecture and training details}

\subsection{Architecture}
In the following, $d_z = d_s = d_h = 512$ and $d_\psi = 32$.

\paragraph{Encoder} The embedding function for each input/output pair is
$$h_\phi(x_i, y_i): (d_x + d_y) \xrightarrow{\text{fc+relu}} d_h \xrightarrow{\text{fc+relu}} d_h \xrightarrow{\text{fc}} d_s\;.$$
For SIVI models, the rest of the encoder $\rho_\phi$ and $\eta_\phi$ is defined as
\begin{align*}
\rho_\phi(s)&: (d_s + d_\epsilon) \xrightarrow{\text{fc+relu}} d_h \xrightarrow{\text{fc}} d_\psi \\
\eta_\phi(s, \psi)&: (d_s + d_\psi) \xrightarrow{\text{fc+relu}} d_h \xrightarrow{\text{fc}} 2*d_z
\end{align*}
where $\epsilon \sim \mathcal{N}(0, I)$ and $\epsilon \in \R^{d_\epsilon}$.
The output of $\eta_\phi$ is then split into two $d_z$-dimensional vectors $\mu_z$ and $\sigma'_z$ with
$$q_\phi(z|\vx_\gC, \vy_\gC) = \mathcal{N}(\mu_z, \text{diag}(0.9 + 0.1*\texttt{sigmoid}(\sigma'_z))^2)\;.$$
For the NP models, there is no $\eta_\phi$, and $\rho_\phi$ is defined as
$$\rho_\phi(s): d_s \xrightarrow{\text{fc+relu}} d_h \xrightarrow{\text{fc}} 2*d_z$$
where the output is split into two vectors $d_z$-dimensional vectors like in SIVI.

\paragraph{Decoder} $g_\theta(x'_j, z): d_x + d_z \underbrace{\xrightarrow{\text{fc+relu}} d_h}_{\text{4 times}} \xrightarrow{\text{fc}} 2*d_y$, where the output is split into two $d_y$-dimensional vectors $\mu_y$ and $\sigma'_y$, and
$$q(y'_j|x'_j, z) = \mathcal{N}(\mu_y, \text{diag}(0.9 + 0.1*\texttt{softplus}(\sigma'_y))^2)\;.$$
For the models with a fixed observation variance, the output of $g_\theta$ is only the vector $\mu_y$ and $q(y'_j|x'_j, z) = \mathcal{N}(\mu_y, 0.2^2 *  \mI_{d_y})$.

ANP model is implemented with the same specifications as above and the other components (deterministic path and attention) is the same as \cite{kim2018attentive}.

\subsection{Training}
All the models were trained using Adam optimizer and a batch size of 16 for 100 epochs. Learning rate was $5\times10^{-4}$ for NP+avg, NP+max and SIVI+max, and $5\times10^{-5}$ for ANP. For SIVI on MNIST, a learning rate scheduler was employed as well that would multiply the learning rate by $0.1$ after 20, 50 and 80 epochs.

The procedure for constructing context sets and target sets from a chosen image in the dataset was as follows. From the image, $n+m'$ pixels, where $n \sim [1, 200)$ and $m' \sim [0, 200)$, were chosen without replacement. The first $n$ pixels constitute the context set, and all $m = n+m'$ pixels were put into the target set.

\subsection{SIVI objective}
As stated in the paper, SIVI bound is a tractable lower bound to the ELBO for hierarchical variational families (c.f. \cref{eq:hierarchical-proposal}). This bound in the context of Neural Processes is defined as
\begin{equation}
    \E_{q_\phi(z, \psi_0|\vx, \vy)}\left[\E_{q_\phi(\psi_{1:K}|\vx, \vy)}\left[ \log \frac{p_\theta(\vy|z, \vx) p_\theta(z)}{\frac{1}{K+1}\sum_{k=0}^K q_\phi(z|\vx, \psi_k)}\right]\right]\label{eq:sivi}
    \leq \text{ELBO}
    \leq \log p_\theta(\vy|\vx)
\end{equation}
where $q_\phi(\psi_{1:K}|\vx) = \prod_{i=1}^K q_\phi(\psi_i|\vx)$ and ELBO is defined as \cref{eq:np_elbo_intial}.

\section{Other variants of Neural Process models} \label{app:other-models}
There are many variants of Neural Processes with different probabilistic modelling assumptions and network architectures. We have attempted to be as clear and fair as possible in generating the qualitative results in the main text. In this section we clarify which specific architectures were considered and why. Moreover, to make the results comparable with other publications in the literature, we include qualitative results for other popular architecture choices not considered in the main text.

\begin{figure}[tbp]
    \centering
    \begin{subfigure}[b]{0.49\linewidth}
        \centering
        \includegraphics[width=\linewidth]{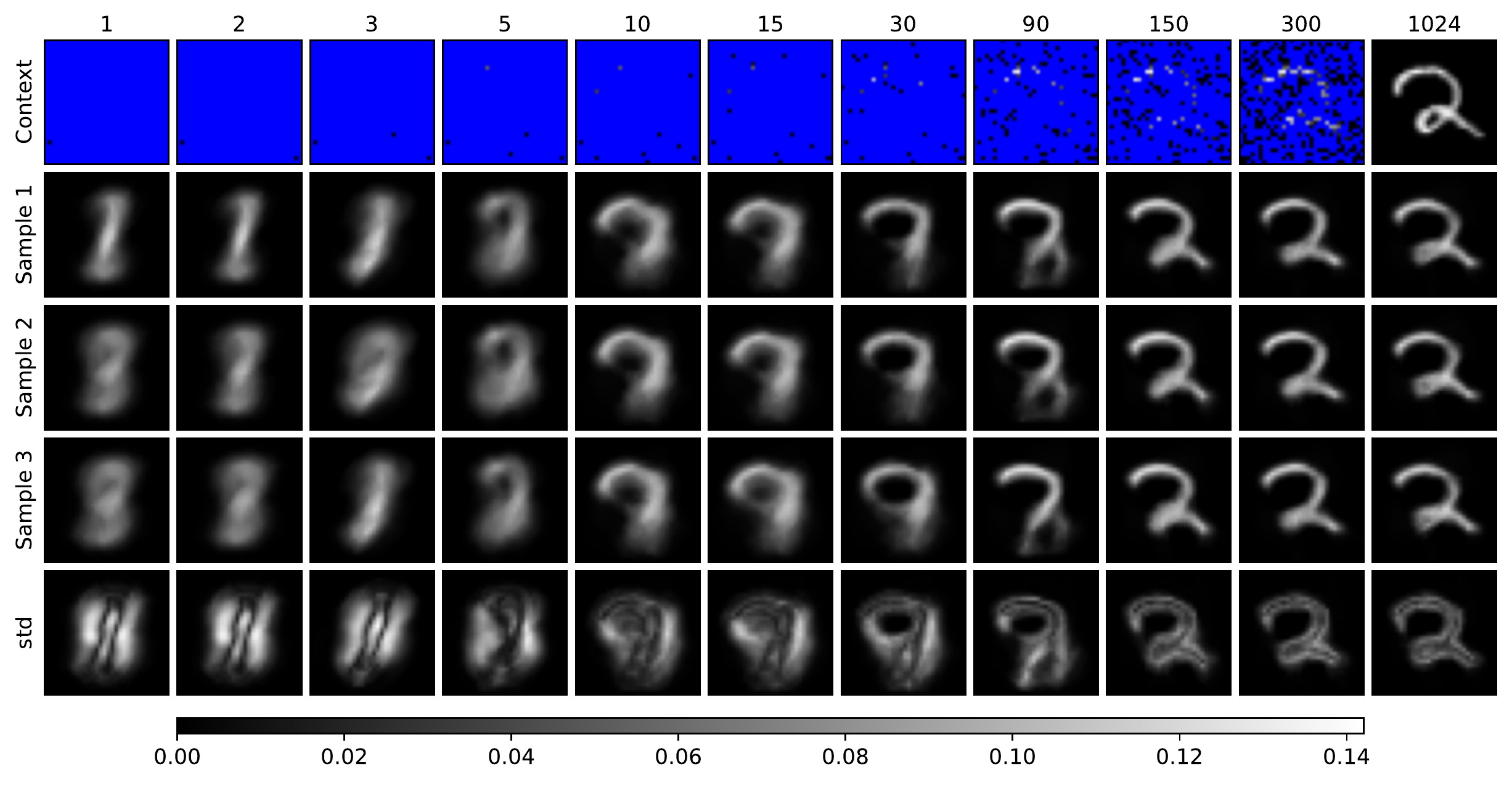}
        \caption{Fixed observation variance}
        \label{fig:app:mnist-fixed-det-avg}
    \end{subfigure}
    \begin{subfigure}[b]{0.49\linewidth}
        \centering
        \includegraphics[width=\linewidth]{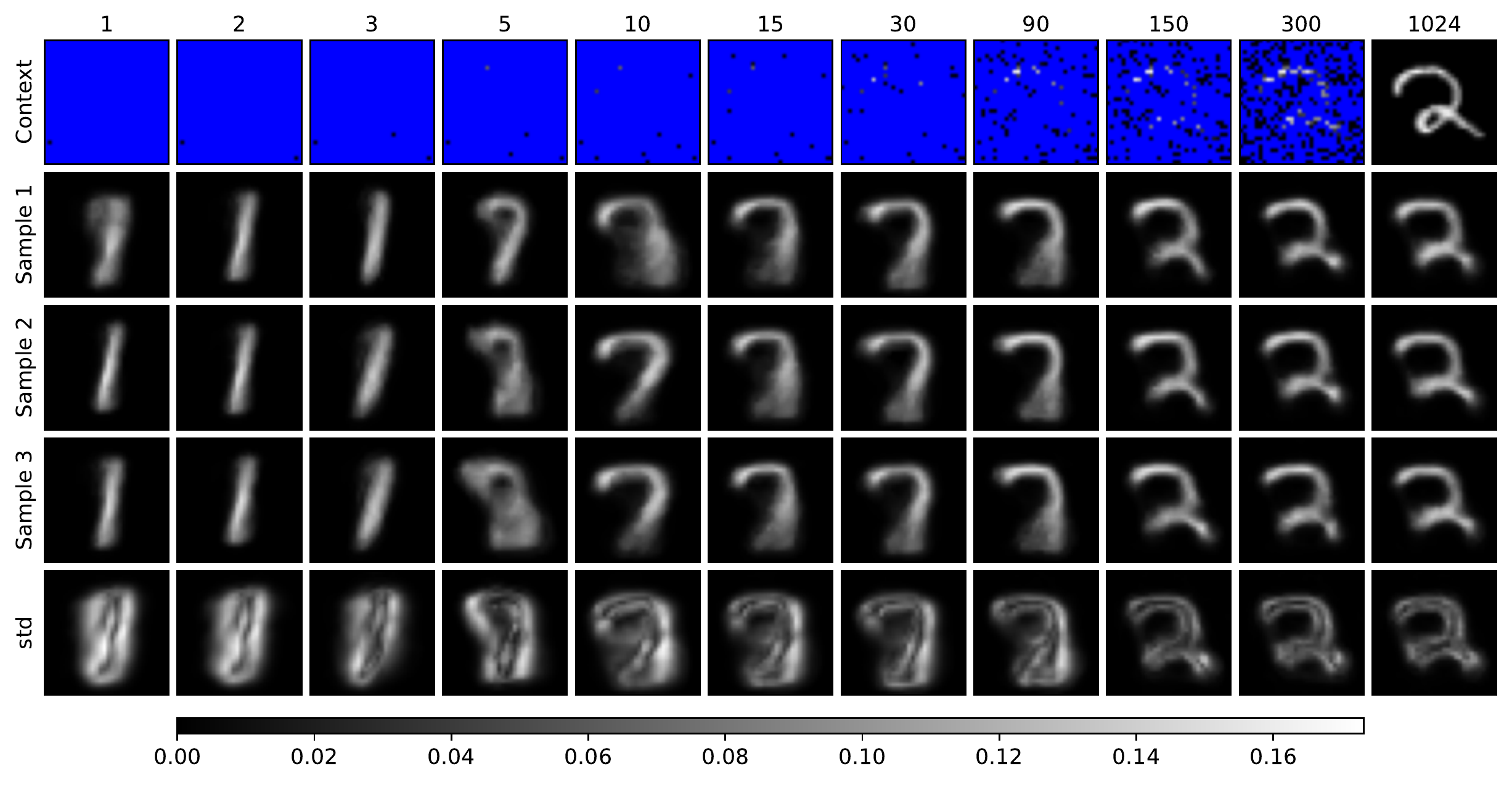}
        \caption{Learned observation variance}
        \label{fig:app:mnist-learned-det-avg}
    \end{subfigure}
    \hfill
    \begin{subfigure}[b]{0.49\linewidth}
        \centering
        \includegraphics[width=\linewidth]{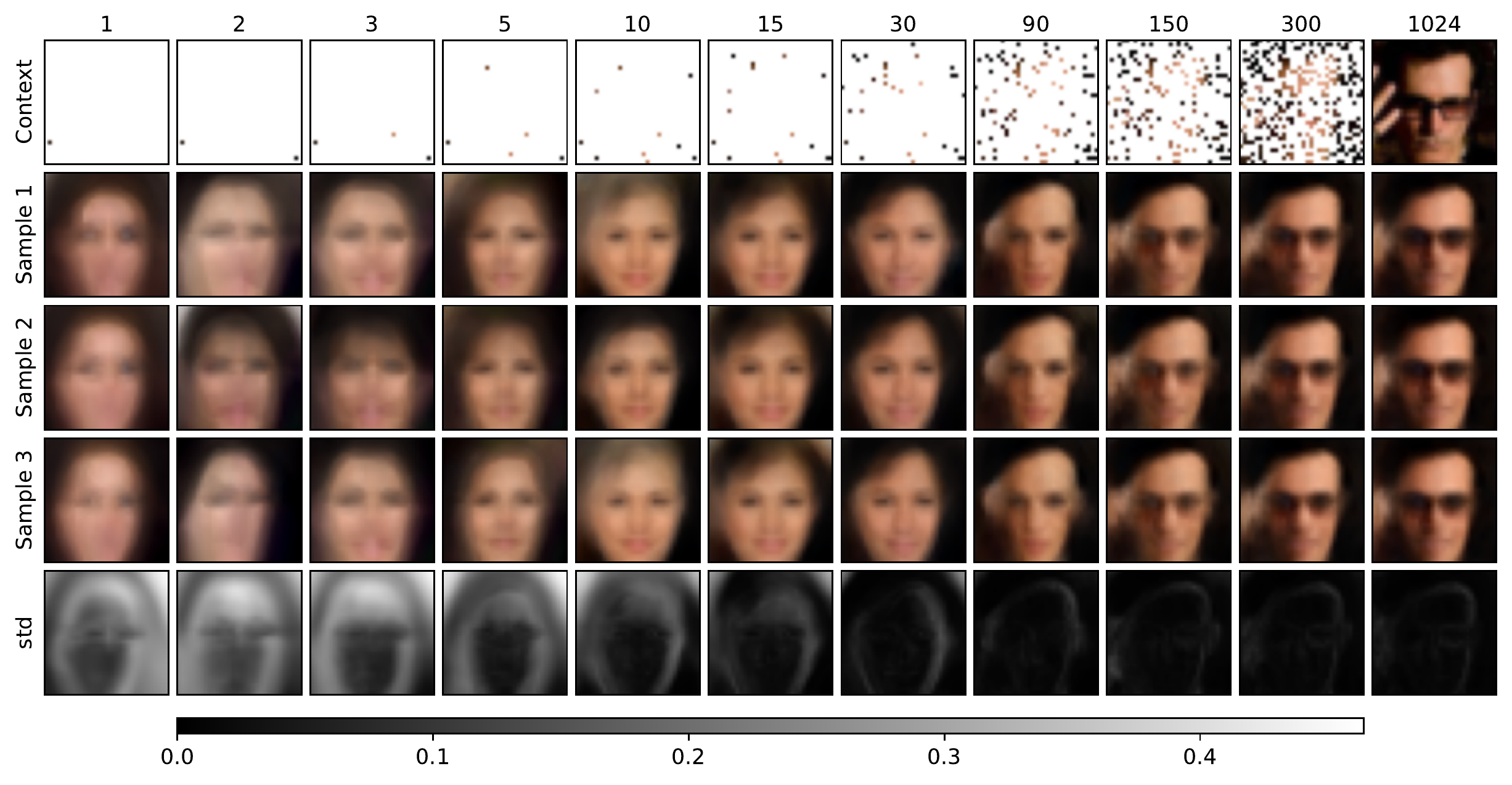}
        \caption{Fixed observation variance}
        \label{fig:app:celeba-fixed-det-avg}
    \end{subfigure}
    \begin{subfigure}[b]{0.49\linewidth}
        \centering
        \includegraphics[width=\linewidth]{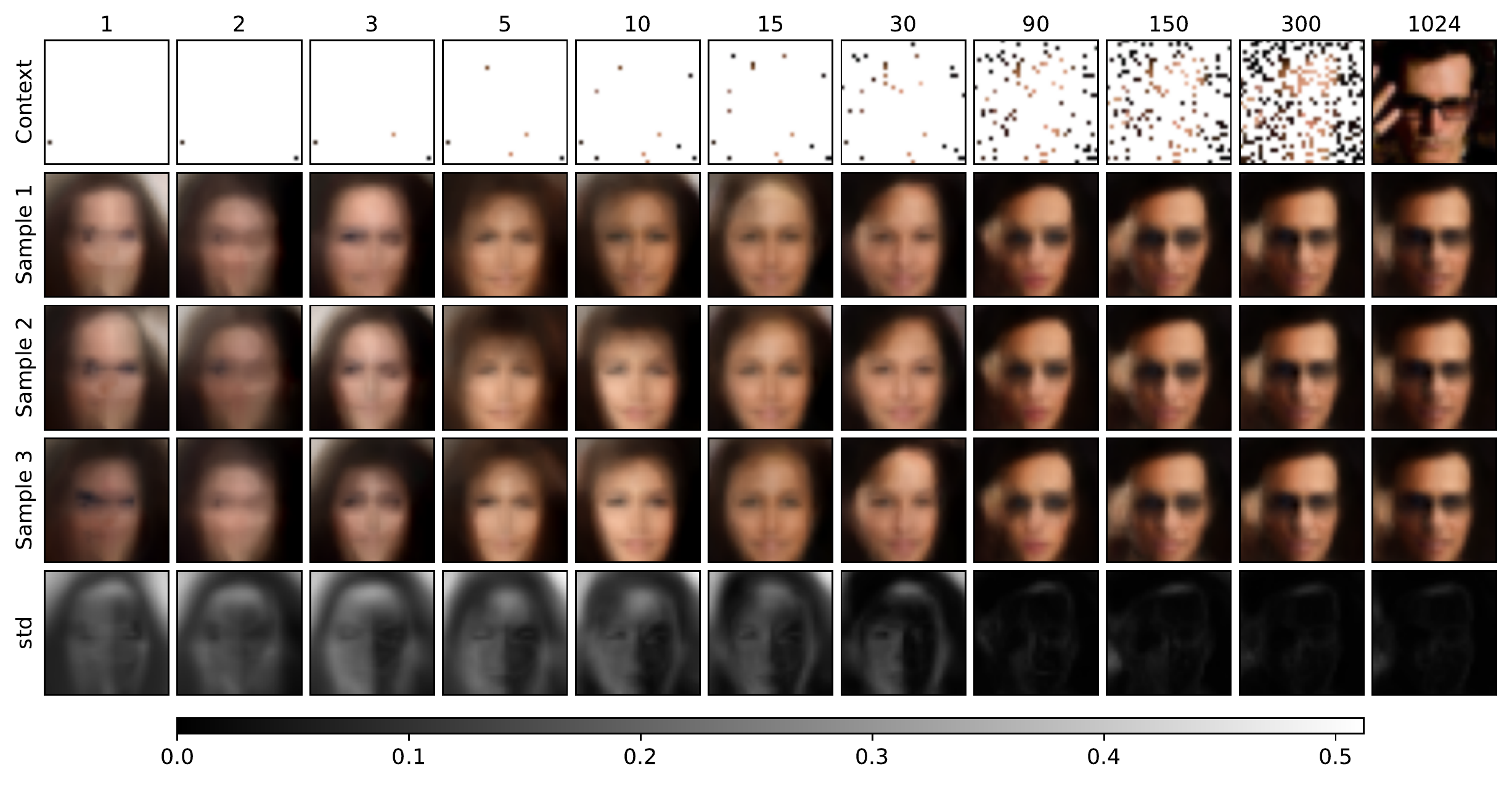}
        \caption{Learned observation variance}
        \label{fig:app:celeba-learned-det-avg}
    \end{subfigure}
    \caption{Qualitative results of NP+avg with deterministic path on (top) MNIST and (bottom) CelebA datasets. These plots show poor sample diversity from the model irrespective of whether the observation variance is fixed or learned.}
    \label{fig:app:det}
\end{figure}

\begin{figure}[tbp]
    \centering
    \begin{subfigure}[b]{0.49\linewidth}
        \centering
        \includegraphics[width=\linewidth]{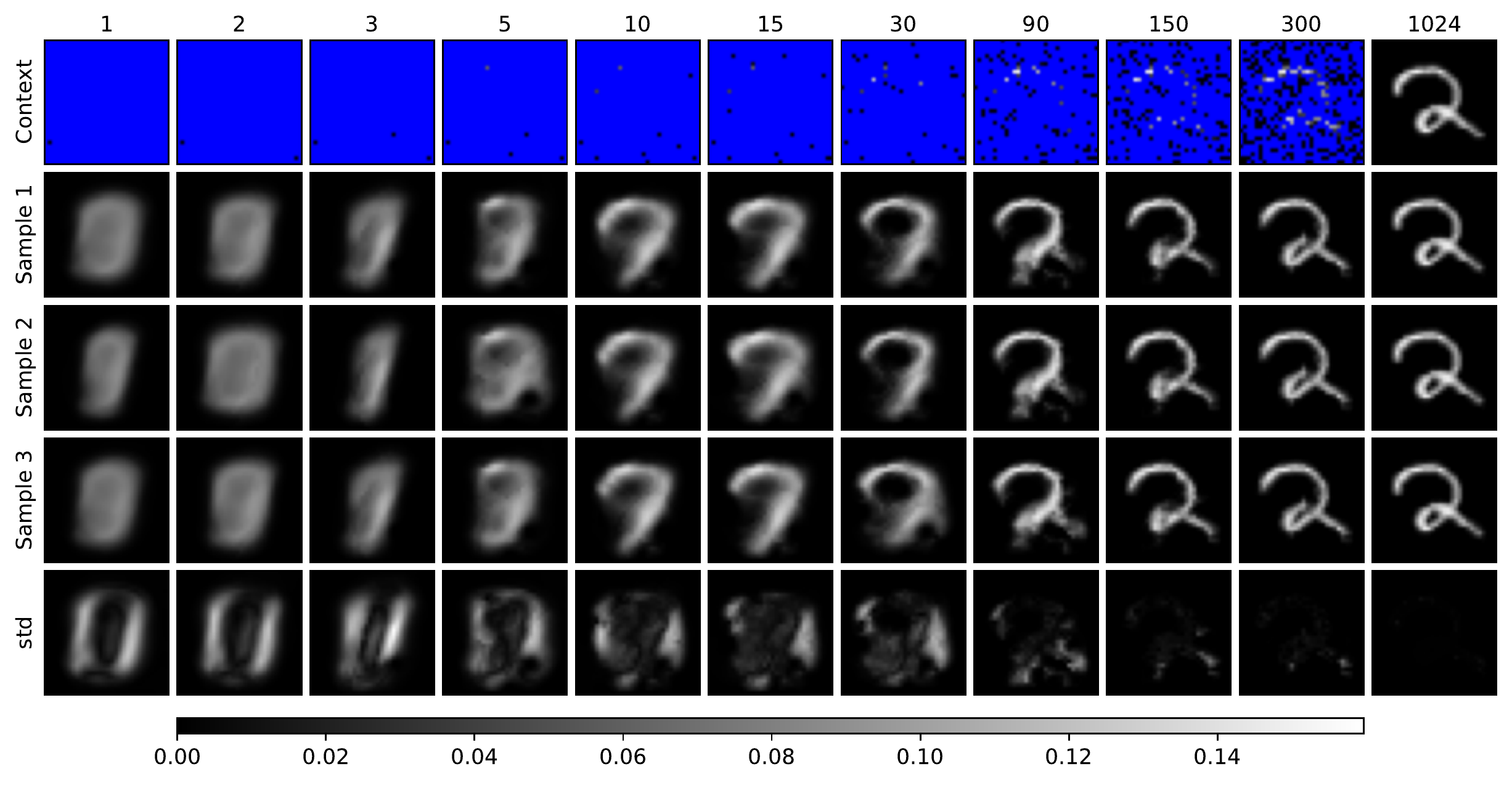}
        \caption{MNIST}
        \label{fig:app:mnist-learned-anp-avg}
    \end{subfigure}
    \begin{subfigure}[b]{0.49\linewidth}
        \centering
        \includegraphics[width=\linewidth]{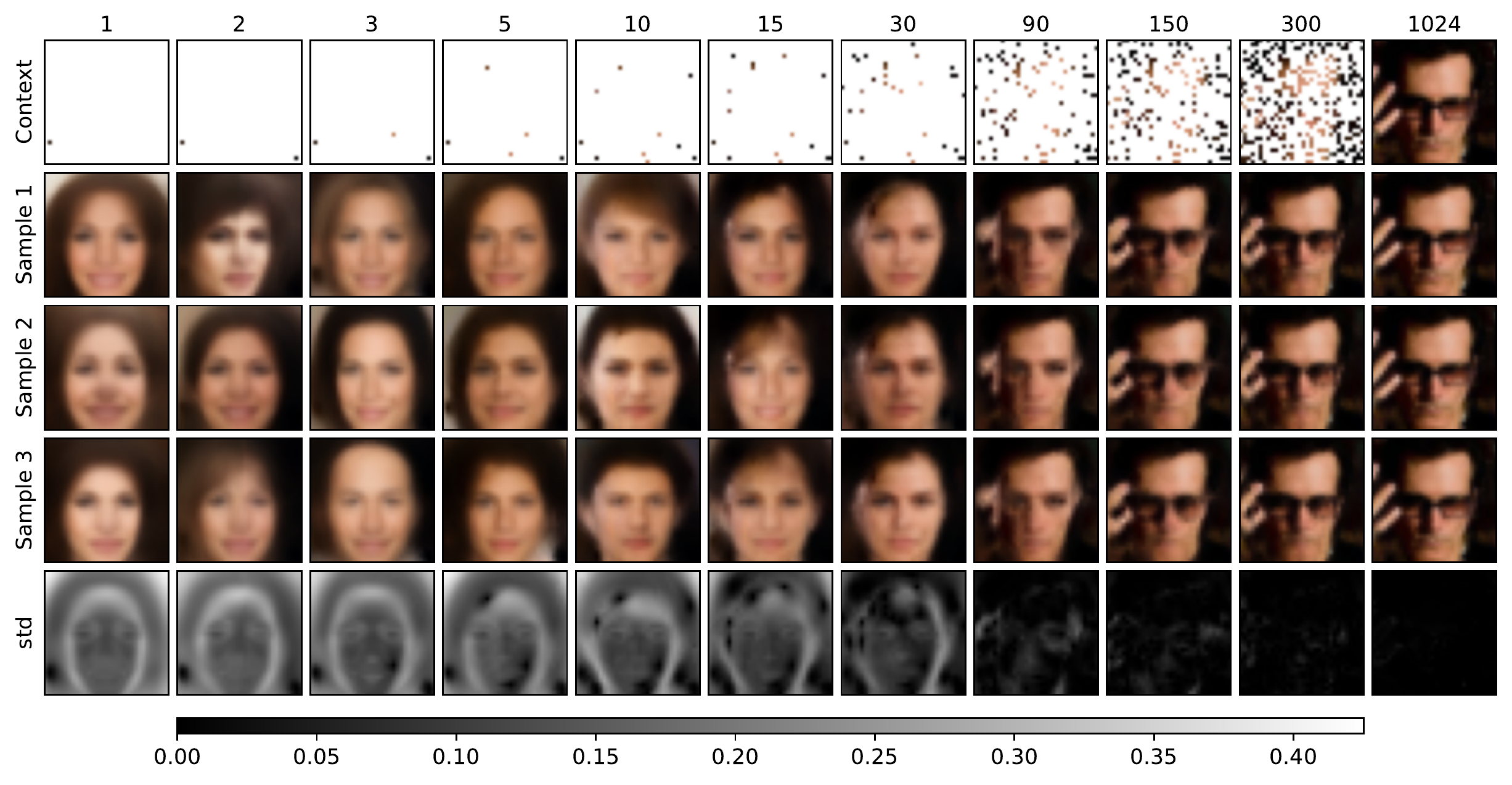}
        \caption{CelebA}
        \label{fig:app:celeba-learned-anp-avg}
    \end{subfigure}
    \caption{Qualitative results of ANP model with learned observation variance. Comparing (\subref{fig:app:mnist-learned-anp-avg}) with \cref{fig:exp:mnist-anp-avg} shows that learning the observation variance hurts sample diversity. It is not as easy to compare sample diversity for CelebA (see (\subref{fig:app:celeba-learned-anp-avg}) and \cref{fig:exp:celeba-anp-avg}). However, ANP in general performs worse than SIVI+max or NP+max on small context sets.}
    \label{fig:app:anp-learned}
\end{figure}

ANPs \citep{kim2018attentive} include a deterministic path bypassing $z$ from the encoder to the decoder that is not found in the original NP \citep{garnelo2018neural}.  Our implementation of NP follows the original model without a deterministic path.  In \citet{kim2018attentive}, a NP without attention but with a deterministic path was considered. \cref{fig:app:det} shows that adding a deterministic path to NP generally hurts sample diversity in small context sizes.  An additional complicating factor is whether $g_\theta$ produces just the mean or both the mean and the variance of the likelihood function.  In line with previous results \citep{le2018empirical}, we found that if $g_\theta$ is trained to produce the observation variance of $p_\theta(y'_i|z, x'_i)$, then models with a deterministic path (including ANP) tend to end up with a large observation variance and a low-variance task-embedding posterior $q_\phi(z|\vx_\gC, \vy_\gC)$, leading to poorly calibrated uncertainty and low sample diversity. Therefore, to be as fair as possible to ANP models, its results in the main text correspond to a model trained with a fixed observation variance, whereas NP and NP+SIVI results are reported with learned observation variance. \cref{fig:app:anp-learned} shows the results for ANP with learned observation variance.

\section{Fig. \ref{fig:exp:max-n} Experimental Details} \label{app:max-n}
\begin{figure}[tbp]
    \centering
    \begin{subfigure}[b]{0.49\linewidth}
        \centering
        \includegraphics[width=\linewidth]{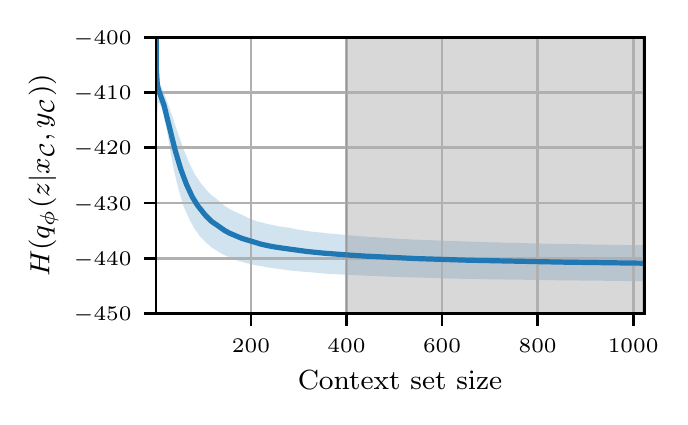}
        \caption{Variational posterior entropy}
        \label{fig:app:avg-posterior-contraction}
    \end{subfigure}
    \begin{subfigure}[b]{0.49\linewidth}
        \centering
        \includegraphics[width=\linewidth]{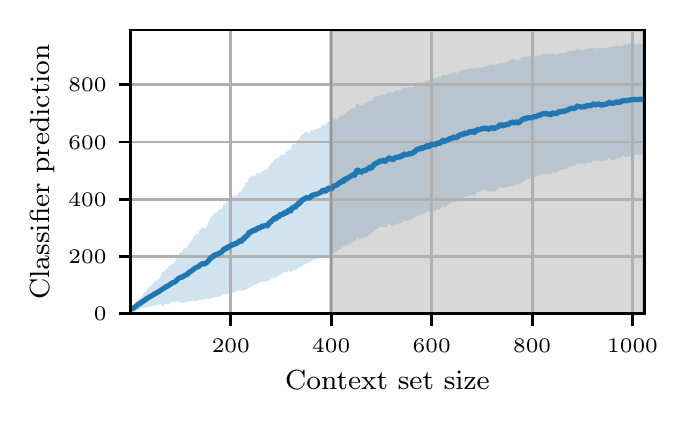}
        \caption{Classifier prediction}
        \label{fig:app:avg-n-classifier}
    \end{subfigure}
    \caption{Posterior contraction of $q_\phi(z|\vx_\gC, \vy_\gC)$ in a NP+mean pooling model.}
    \label{fig:app:avg-n}
\end{figure}
\cref{fig:exp:max-posterior-contraction} shows entropy of the variational posterior, i.e., $q_\phi(z|\vx_\gC, \vy_\gC)$, versus context set size ($n$) for a growing context set with i.i.d. items. The plot shows an aggregation over 1000 runs of this procedure, each with a different ground truth image. The experiment verifies that the learned NP posterior follows the classical Bayesian inference results and, more interestingly, the posterior contraction even generalizes to context sets larger than the context sets seen during training. The plot was generated by an NP+max model trained on MNIST, but the observed behavior is not specific to it. We see the same behavior when average pooling is used for the CelebA dataset.

The experiment suggests that even though the context dataset is represented through an aggregated embedding that does not explicitly embed $n$, the training objective forces the networks and the embedding space to retain information about $n$. We validate this by training a classifier to predict $n$ given the learned embeddings $s_\gC$. \cref{fig:exp:max-n-classifier} shows the classifier performance on a held-out test set and shows a strong correlation between embeddings $s_\gC$ and context set sizes.

\cref{fig:app:avg-n} shows the same behavior with mean pooling. The plot is generated by a NP+mean model trained on MNIST dataset.

\section{Max pooled embeddings and posterior entropy}
\label{app:max-pool}
\begin{figure}[tbp]
    \centering
    \begin{subfigure}[b]{0.49\linewidth}
        \centering
        \includegraphics[width=\linewidth]{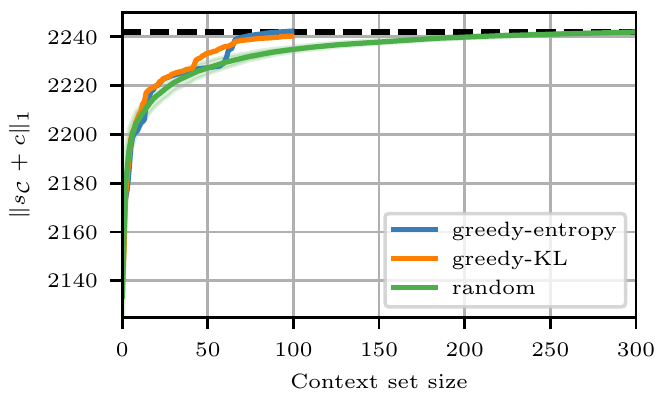}
        \caption{}
        \label{fig:max-pool-contract-norm}
    \end{subfigure}
    \begin{subfigure}[b]{0.49\linewidth}
        \centering
        \includegraphics[width=\linewidth]{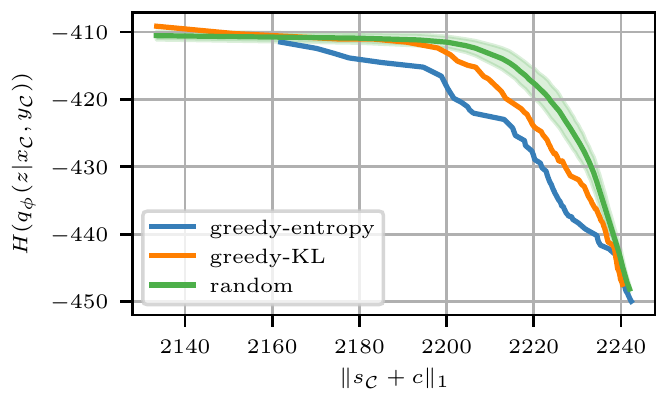}
        \caption{}
        \label{fig:max-pool-contract-entropy}
    \end{subfigure}
    \caption{(a) Norm of pooled embedding, $||s_\gC||_1$, versus context size for three different methods of context set generation. Note that the embeddings are shifted so that the minimum embedding value in each dimension is 0. (b) Posterior entropy versus norm of (shifted) pooled embedding. Observe that the norm of the embedding is strictly increasing in context size, with large increases when the context is small; and that the posterior entropy is decreasing as a function of the norm of the embedding.}
    \label{fig:max-pool-contract}
\end{figure}

As discussed in the main text, a  NP model with max pooling exhibits posterior contraction by learning a $\rho_\phi$ such that the posterior entropy is a decreasing function in all dimensions of embedding space. To illustrate, \cref{fig:max-pool-contract} shows $||s_\gC||_1$ vs context size (increasing), and the posterior entropy versus $||s_\gC||_1$ (decreasing).

\section{Computing Test Data Log Likelihoods}
The test data (normalized) log likelihoods $\frac{1}{|\gT|}\log p_\theta(\vy_\gT|\vx_\gT, \vx_\gC, \vy_\gC)$ are computed and averaged over context/target sets sampled from held-out test sets. Context sets and target sets are \textit{disjoint} (i.e., all the items in target set are unobserved) and have a random size in $[1,200)$. As we do not have a closed form for predictive log-likelihoods, we compute the following IWAE-like lower bound \citep{burda2016importance} instead with K=1000.
\begin{align}
    \log p_\theta(\vy_\gT|\vx_\gT, \vx_\gC, \vy_\gC)
    &\geq \E_{q_\phi(z_{1:K}|\vx_\gC, \vy_\gC)} \log \frac{1}{K} \sum_{k=1}^K \frac{p_\theta(\vy_\gT|\vx_\gT, z) p_\theta(z|\vx_\gC, \vy_\gC)}{q_\phi(z_k|\vx_\gC, \vy_\gC)}\\
    &\approx \E_{q_\phi(z_{1:K}|\vx_\gC, \vy_\gC)} \log \frac{1}{K} \sum_{k=1}^K \frac{p_\theta(\vy_\gT|\vx_\gT, z_k) q_\phi(z|\vx_\gC, \vy_\gC)}{q_\phi(z_k|\vx_\gC, \vy_\gC)}\\
    &= \E_{q_\phi(z_{1:K}|\vx_\gC, \vy_\gC)} \log \frac{1}{K} \sum_{k=1}^K p_\theta(\vy_\gT|\vx_\gT, z_k)\;.
\end{align}

\section{Computing Inception Scores in our experiments}
\subsection{Definition}
Inception score is defined in a way such that a high score requires the individual samples to be classifiable with high confidence and, at the same time, the marginal class distribution of samples to be diverse. More formally,
\begin{equation}
    \log \text{IS}
    = \E_{\vx \sim G}\left[ \KL(p(y|\vx) || p(y)) \right]
    = - \E_{\vx \sim G} \left[ H(p(y|\vx)) - H(p(y|\vx), p(y))  \right]
\end{equation}
where $G$ is a generator producing samples $\vx$ and $y$ is the classification labels specified by the classifier.

\subsection{Classifier networks}
As results in the GAN literature suggest that inception score is unreliable when applied to image domains other than ImageNet, we replace inception network with classifiers trained on MNIST and CelebA datasets. The network architecture of both classifiers is ResNet \citep{he2016deep}. The MNIST classifier network is trained to solve the MNIST digit classification task with 10 classes. It is more challenging for CelebA as there is no well-defined set of labelled classes for it. As CelebA images are labelled with 40 attributes, we choose the four attributes of \{Male, Black Hair, Smiling, Young\} and construct a synthetic classification task with $2^4$ classes where each class refers to a configuration of the chosen attributes. The trained models are used in place of inception network to get calibrated scores in our experiments.

\section{MNIST classifier results} \label{app:mnist-classifier}
We examine the diversity of samples generated from each model by classifying them using a MNIST classifier and looking at the distribution of the predictions. The main expectations are that (1) the models have a non-zero probability of generating the ground truth image irrespective of the context set and that (2) the models do not generate digits that are inconsistent with the context set.

\begin{figure}[htbp]
    \centering
    \includegraphics[width=\linewidth]{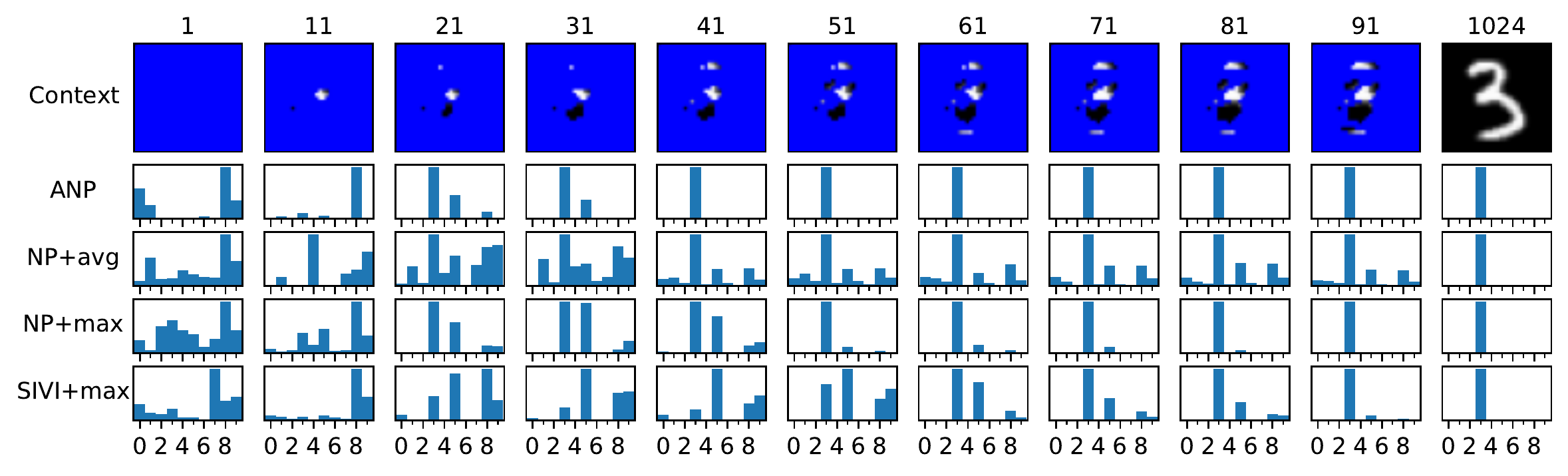}
    \caption{MNIST classification results for a sequence of growing context sets. Each column shows the results for the context set in the first row. Each histogram under a context set shows the prediction distribution of a MNIST classifier for 1000 samples from a model that was conditioned on the context set. The context set sizes are written at the top of each column. The first context set is the top left pixel, treated as an uninformative context set. Each of the following context sets add 10 new pixels that are specifically chosen to eliminate a remaining possible digit (in the order of 0 to 9). Given the digit to eliminate, the 10 chosen pixels are the ones that differ the most in pixel intensity between the mean image of all instances of 3 in the training set and the mean image of all instances of the digit to eliminate.}
    \label{fig:app:mnist-classifier-v1}
\end{figure}

\begin{figure}[htbp]
    \centering
    \includegraphics[width=0.7\linewidth]{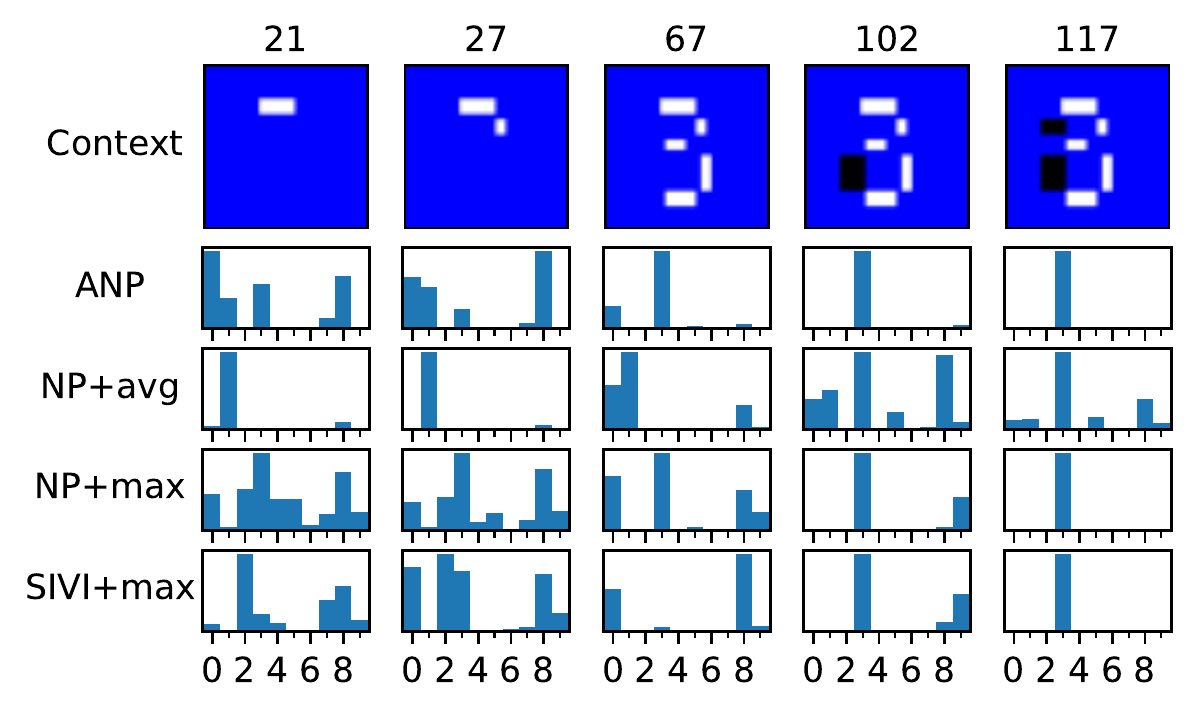}
    \caption{MNIST classification results for a synthetic sequence of context sets.
    The context sets are designed to hint at an image of a 3. The details of how the subplots are organized is the same as \cref{fig:app:mnist-classifier-v1}. All the models except NP+avg have a non-zero probability for the correct digit throughout the process. In terms of the compatibility of the generated digits with the context set, ANP works reasonably well on larger context sets while NP+max and SIVI+max generally outperform the others throughout the process.}
    \label{fig:app:mnist-classifier-v3}
\end{figure}

As the true posterior probability of the digit given a few pixels of its image (the context set) is unknown, we report \cref{fig:app:mnist-classifier-v1,fig:app:mnist-classifier-v3} as a proxy to it. These figures show the results of an experiment where a sequence of growing context sets incrementally reveals an image of a 3 and compare the prediction distribution of generated samples from different models. The final image in \cref{fig:app:mnist-classifier-v1} is chosen from the test set, and the context sets are constructed to eliminate a specific digit with each step. In \cref{fig:app:mnist-classifier-v3}, the final image is synthetic and hand-drawn. The context set in each step is grown by adding new strokes of the digit.

\section{Additional Qualitative Results}
\begin{figure}[tbp]
    \centering
    \begin{subfigure}[b]{0.45\linewidth}
        \centering
        \includegraphics[width=\linewidth]{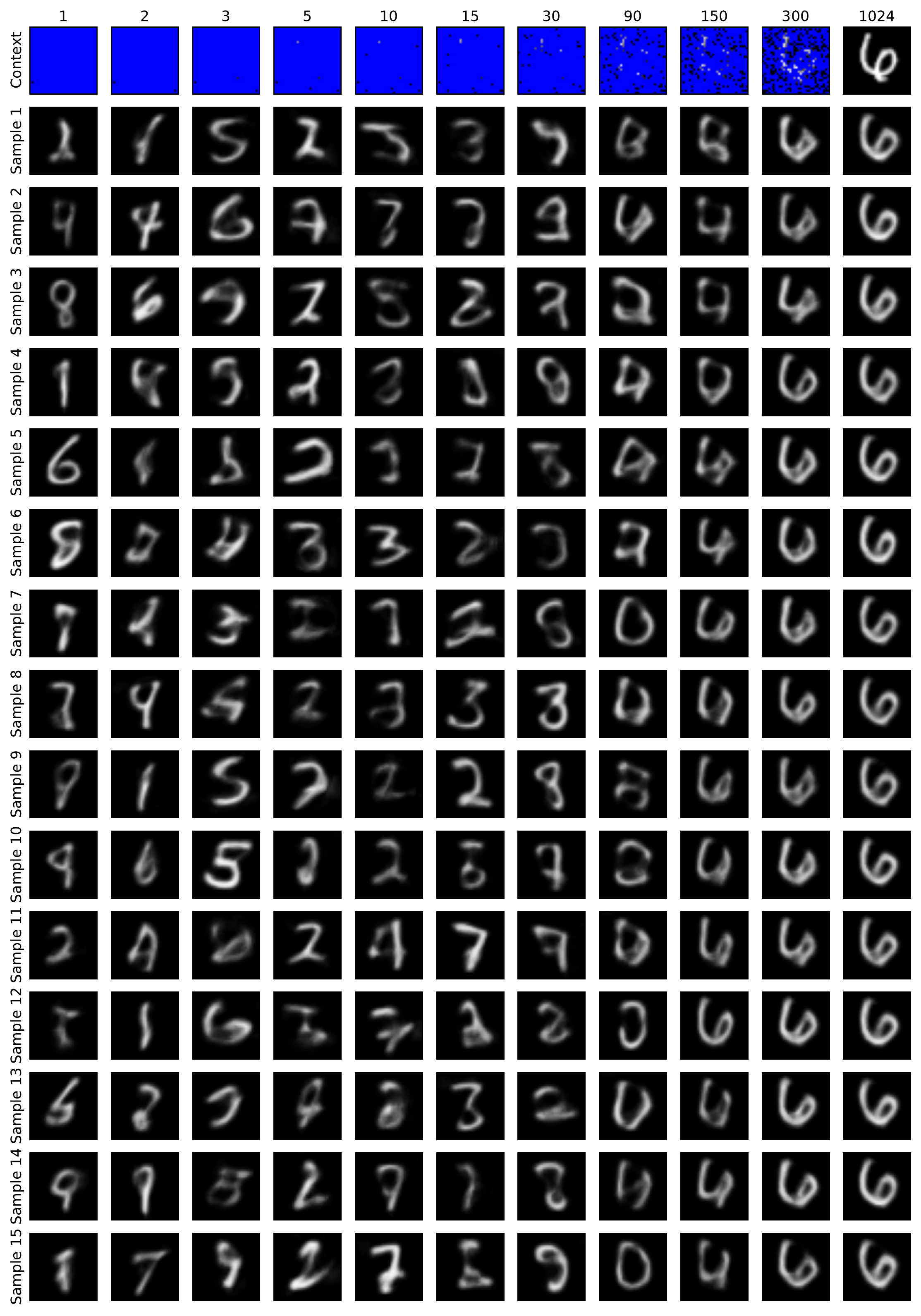}
        \caption{NP objective with average pooling}
    \end{subfigure}
    \begin{subfigure}[b]{0.45\linewidth}
        \centering
        \includegraphics[width=\linewidth]{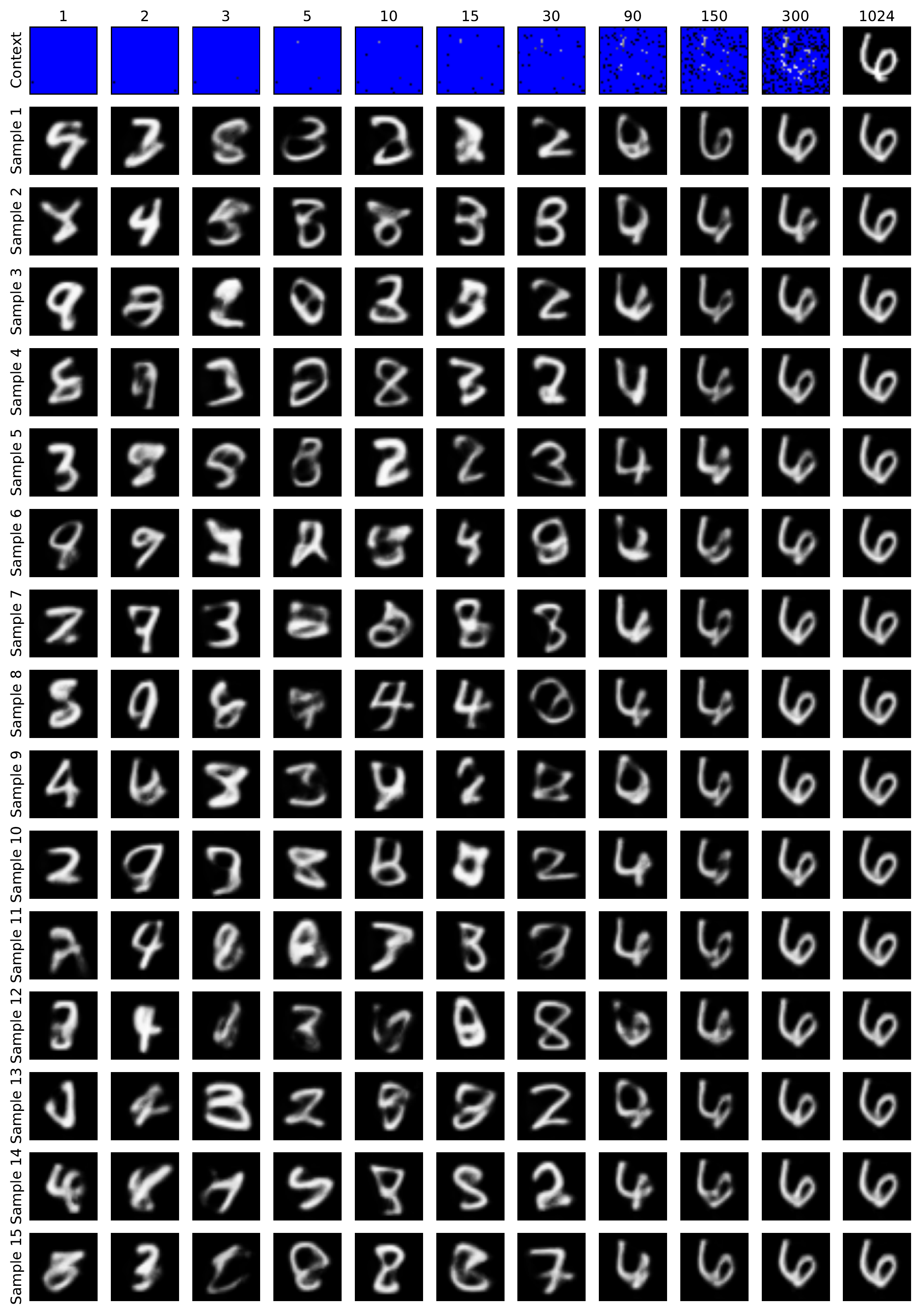}
        \caption{NP objective with max pooling}
    \end{subfigure}
    \hfill
    \begin{subfigure}[b]{0.45\linewidth}
        \centering
        \includegraphics[width=\linewidth]{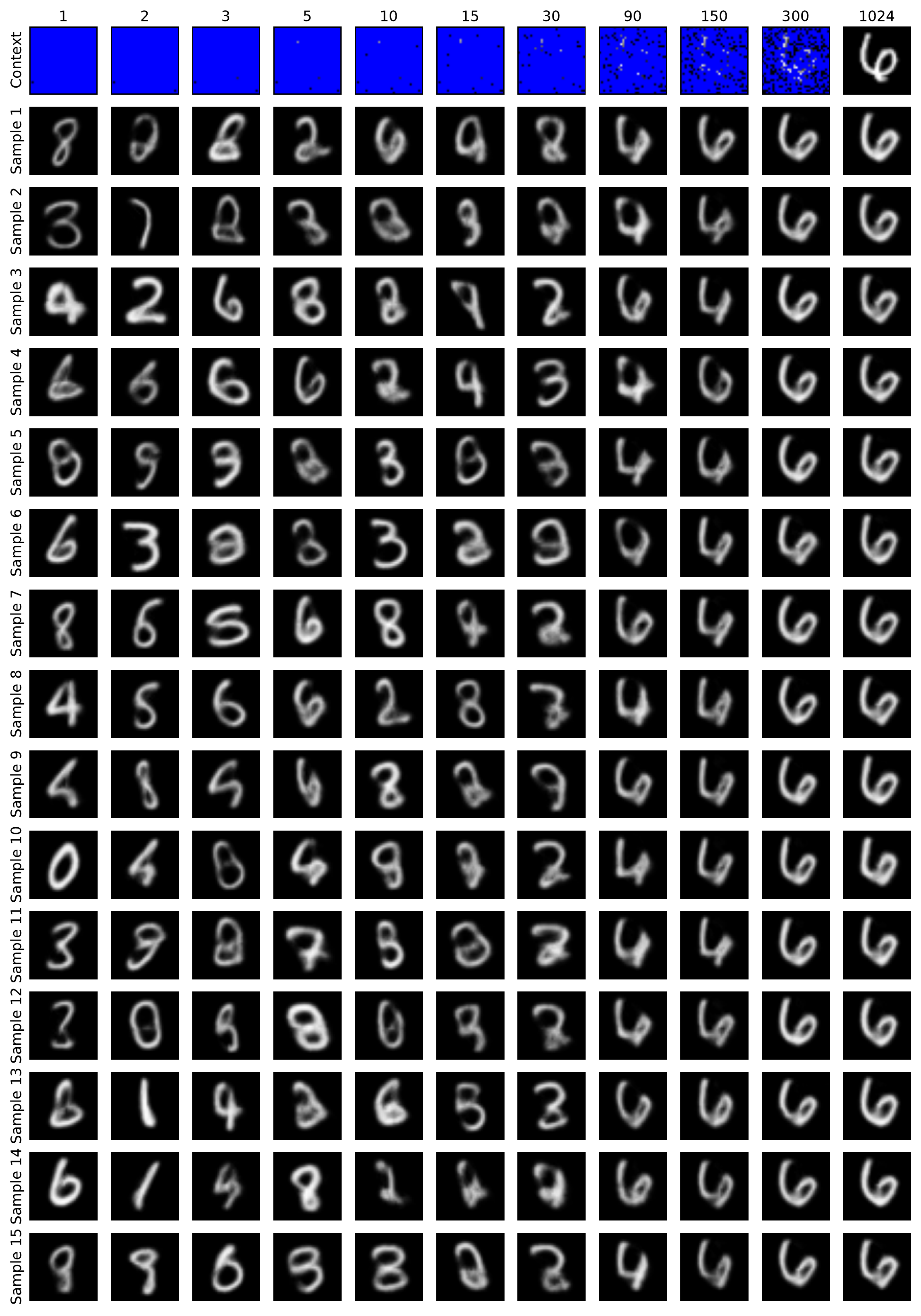}
        \caption{NP+SIVI objective with max pooling}
    \end{subfigure}
    \begin{subfigure}[b]{0.45\linewidth}
        \centering
        \includegraphics[width=\linewidth]{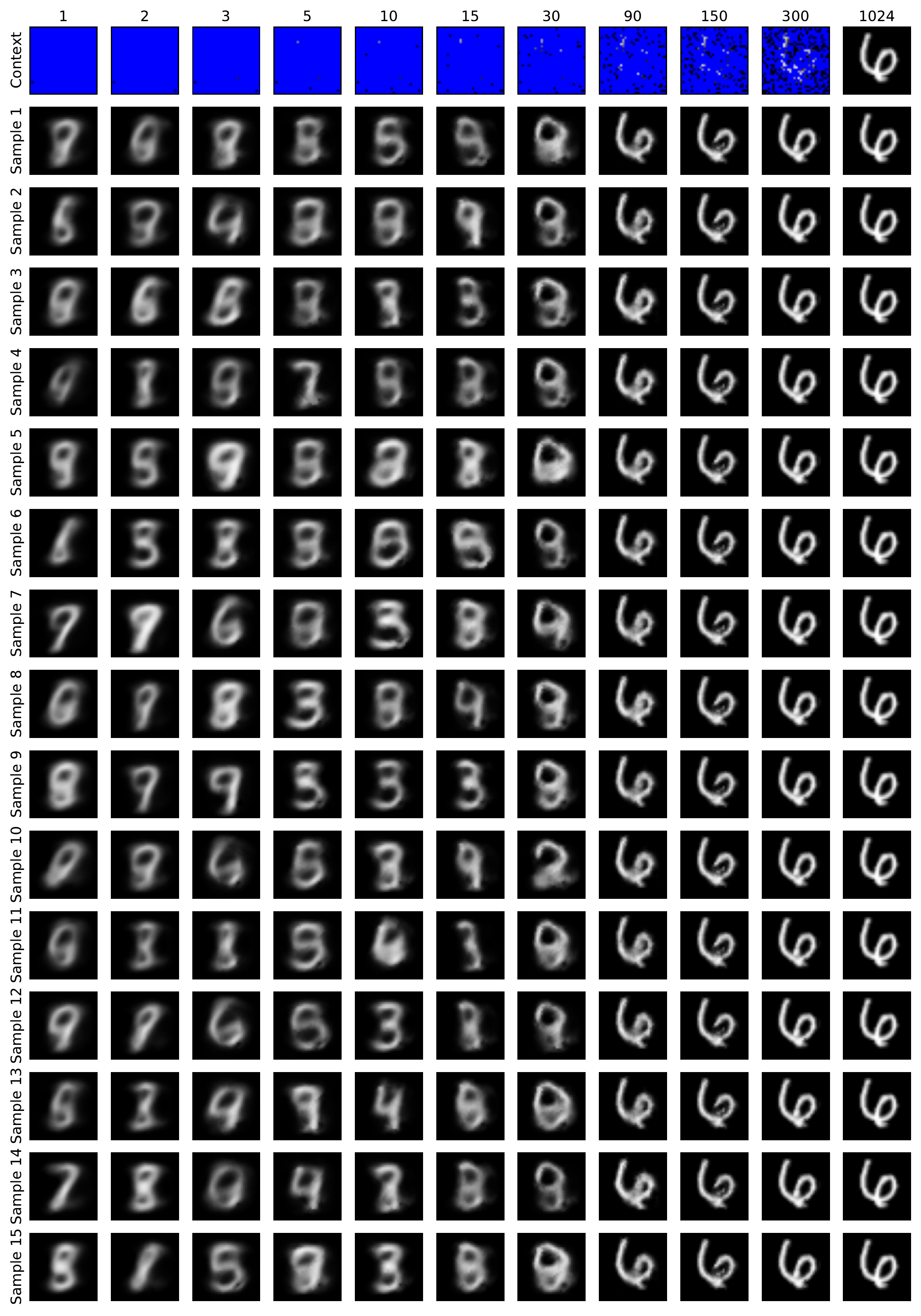}
        \caption{ANP}
    \end{subfigure}
    \caption{Samples from models trained on the MNIST dataset.}
    \label{app:additional-mnist}
\end{figure}

\begin{figure}[tbp]
    \centering
    \begin{subfigure}[b]{0.45\linewidth}
        \centering
        \includegraphics[width=\linewidth]{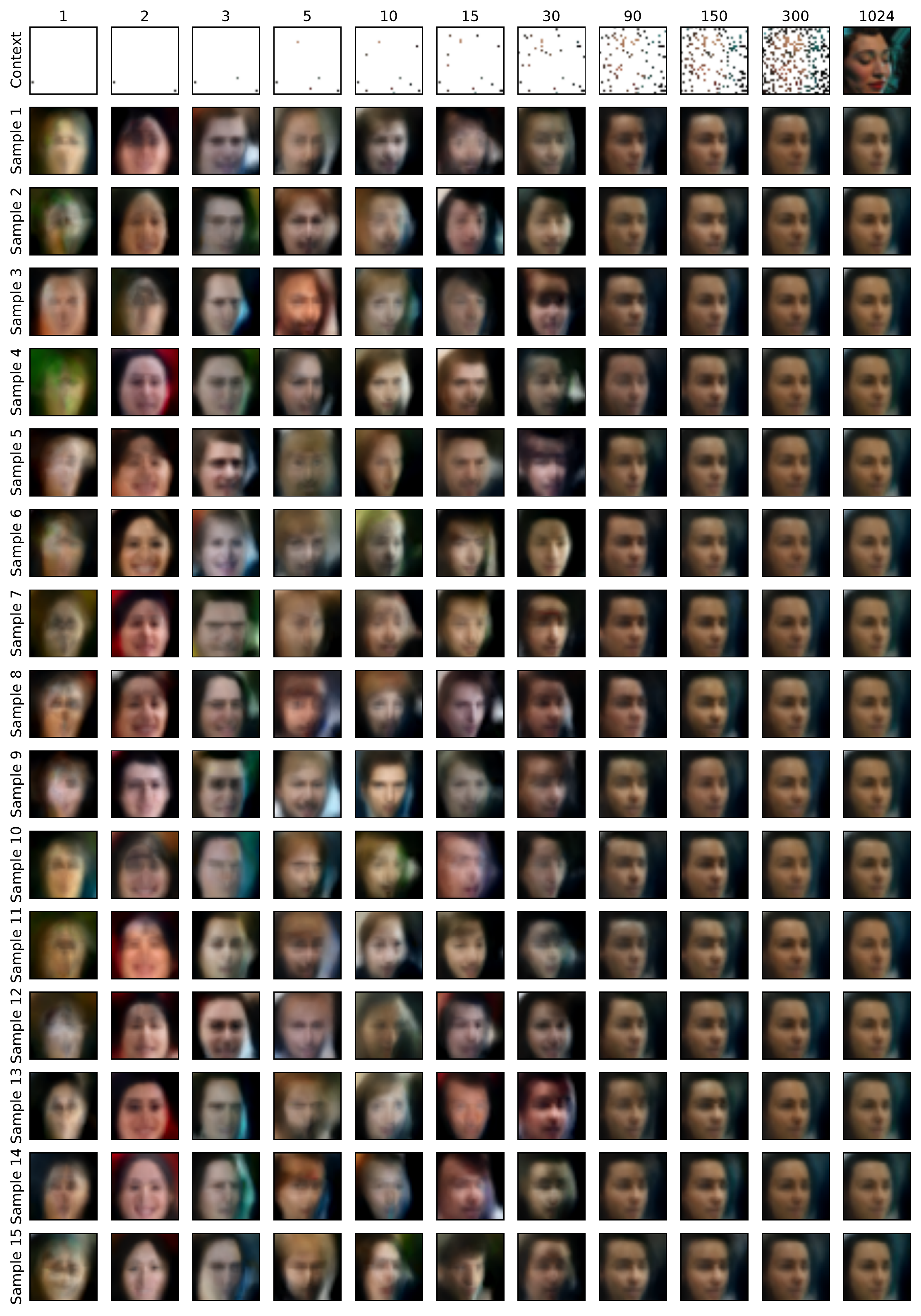}
        \caption{NP objective with average pooling}
    \end{subfigure}
    \begin{subfigure}[b]{0.45\linewidth}
        \centering
        \includegraphics[width=\linewidth]{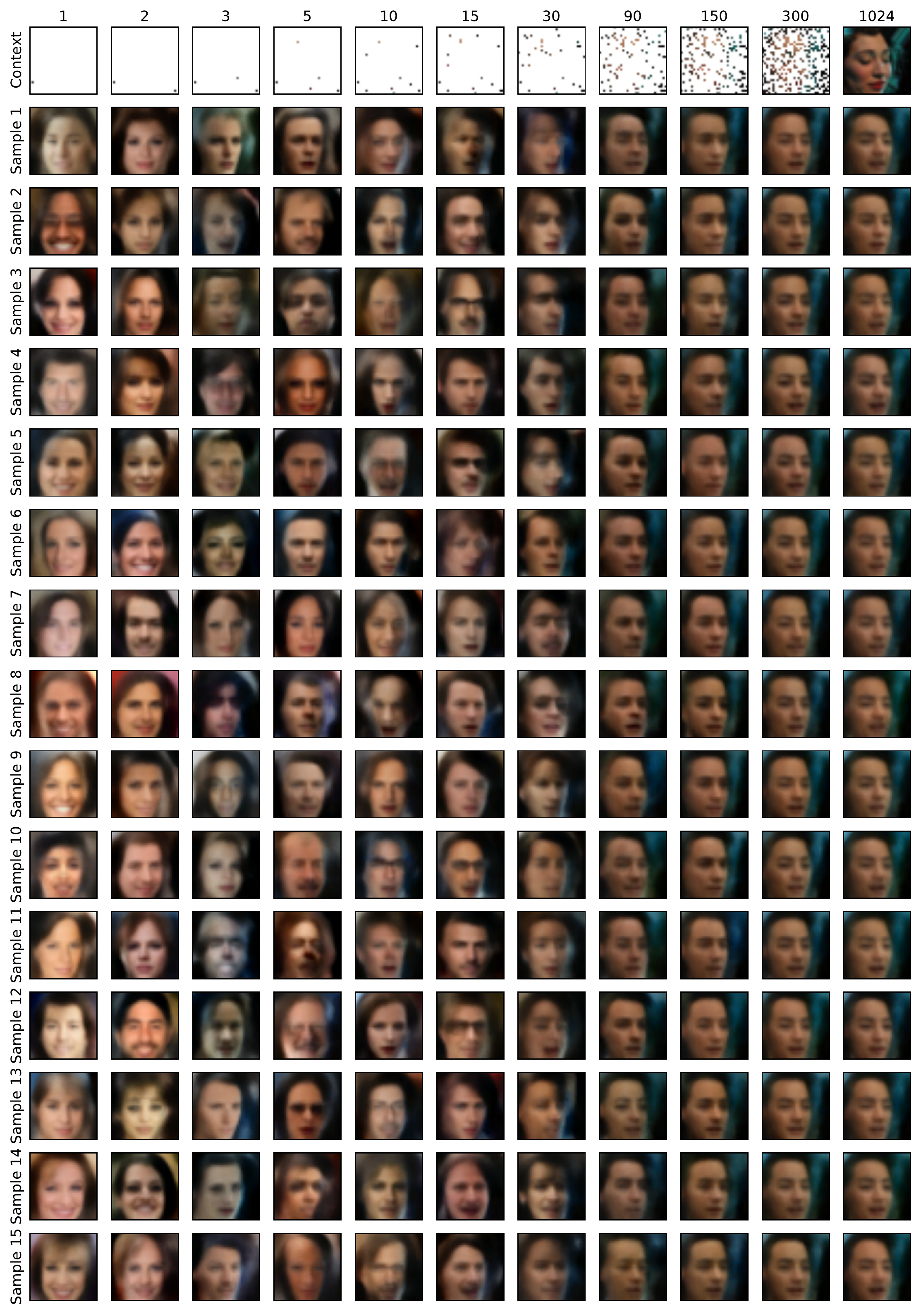}
        \caption{NP objective with max pooling}
    \end{subfigure}
    \hfill
    \begin{subfigure}[b]{0.45\linewidth}
        \centering
        \includegraphics[width=\linewidth]{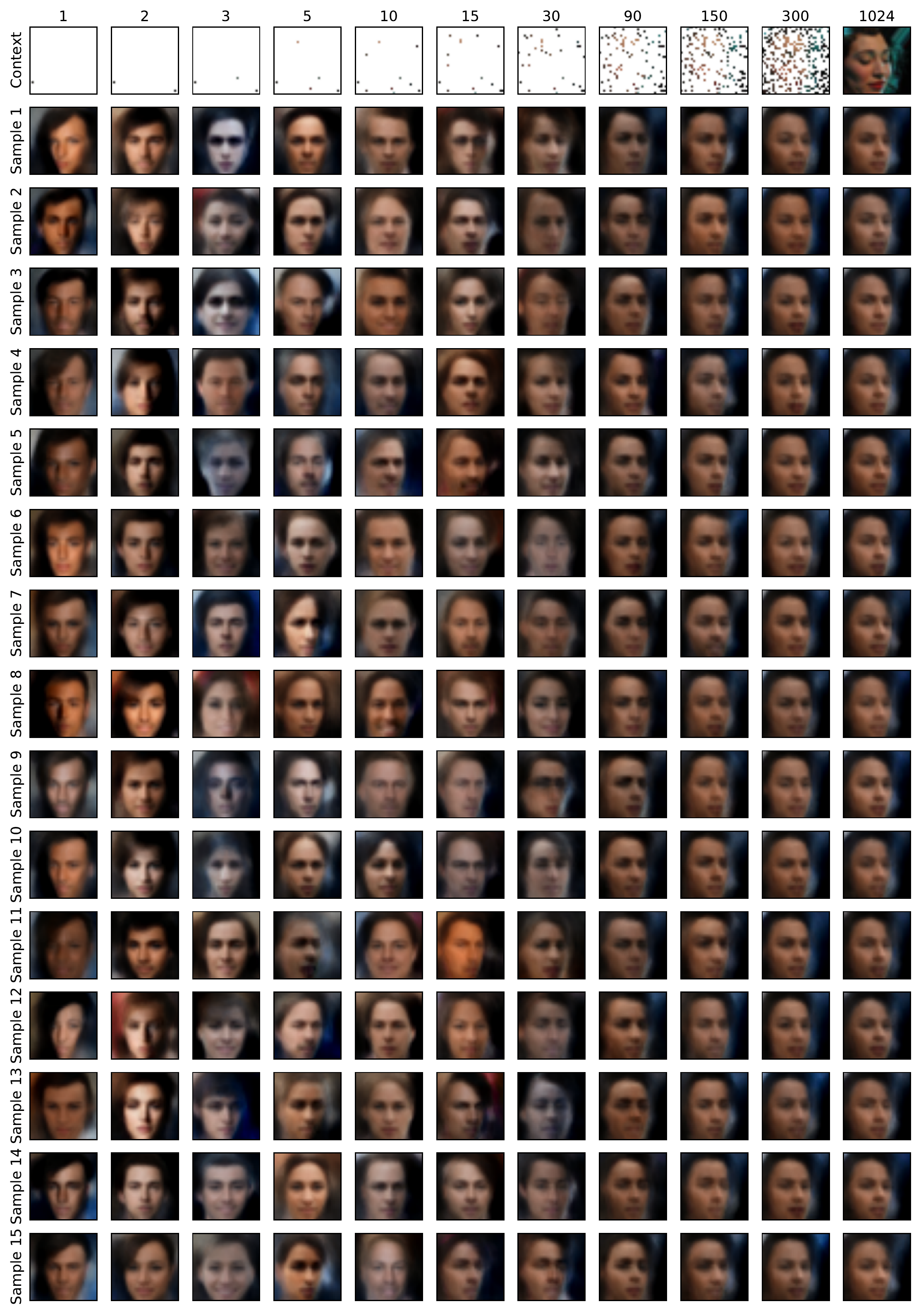}
        \caption{NP+SIVI objective with max pooling}
    \end{subfigure}
    \begin{subfigure}[b]{0.45\linewidth}
        \centering
        \includegraphics[width=\linewidth]{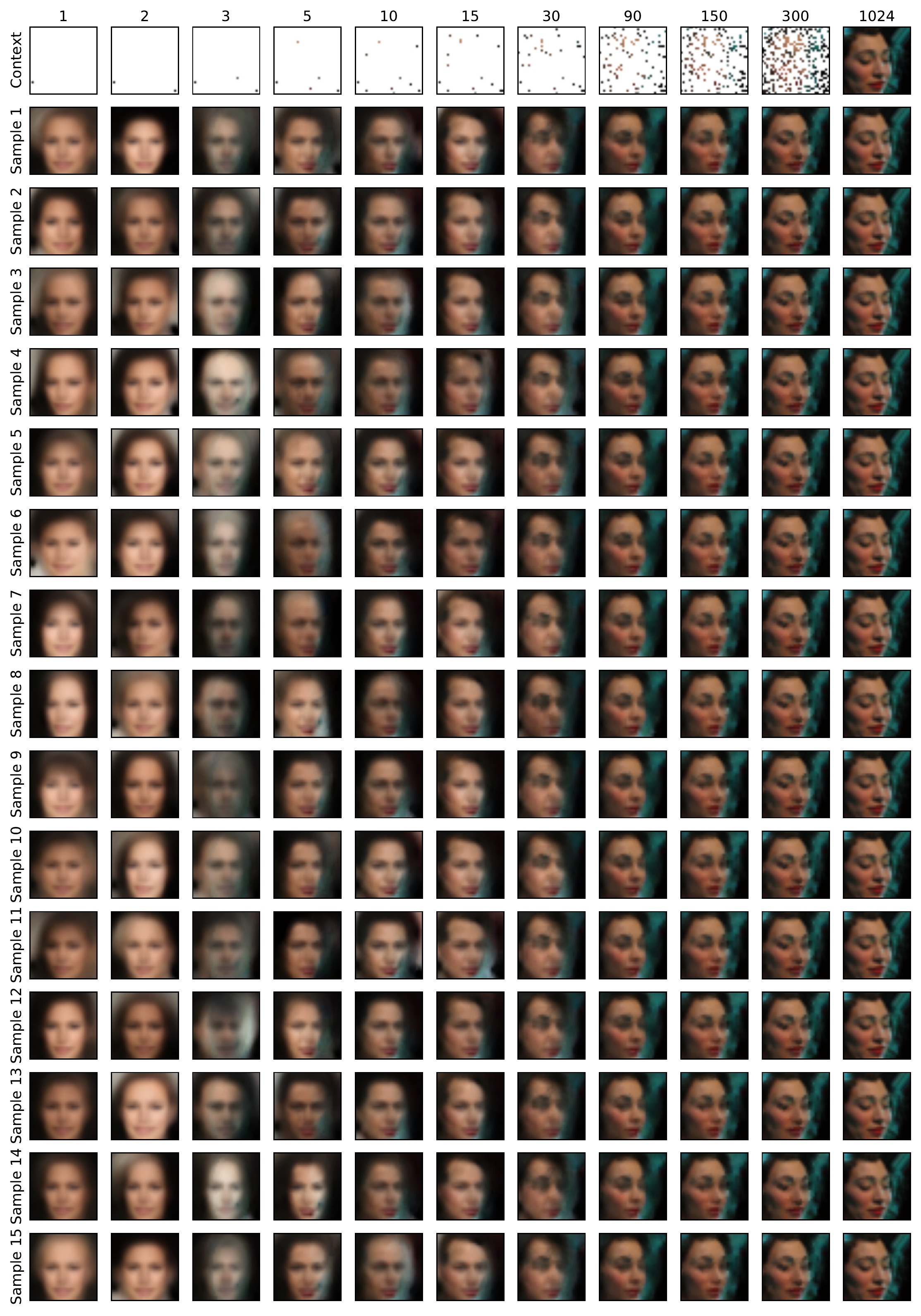}
        \caption{ANP}
    \end{subfigure}
    \caption{Samples from models trained on the CelebA dataset.}
    \label{app:additional-celeba}
\end{figure}

\begin{figure}[tbp]
    \centering
    \begin{subfigure}[b]{0.45\linewidth}
        \centering
        \includegraphics[width=\linewidth]{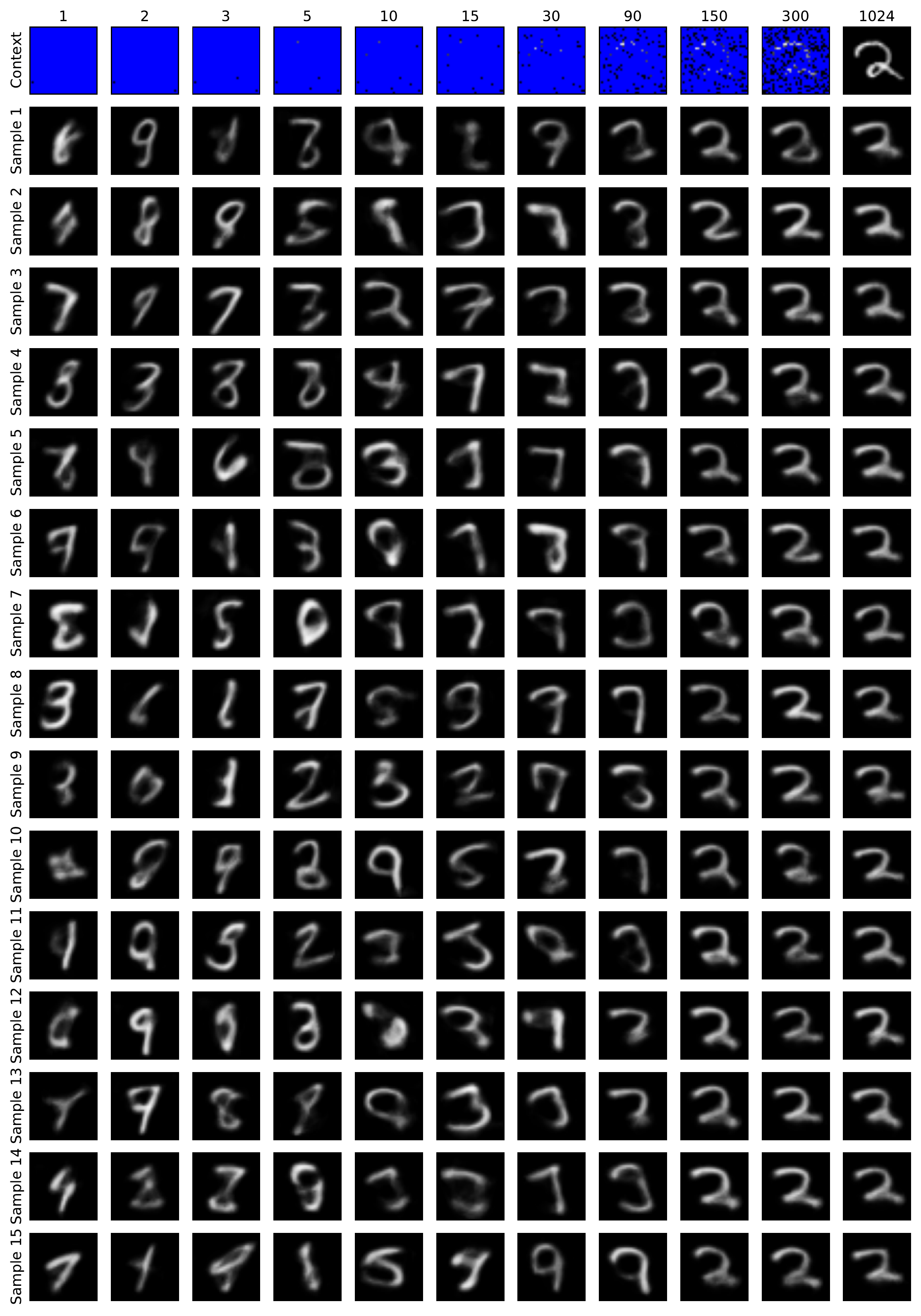}
        \caption{NP objective with average pooling}
    \end{subfigure}
    \begin{subfigure}[b]{0.45\linewidth}
        \centering
        \includegraphics[width=\linewidth]{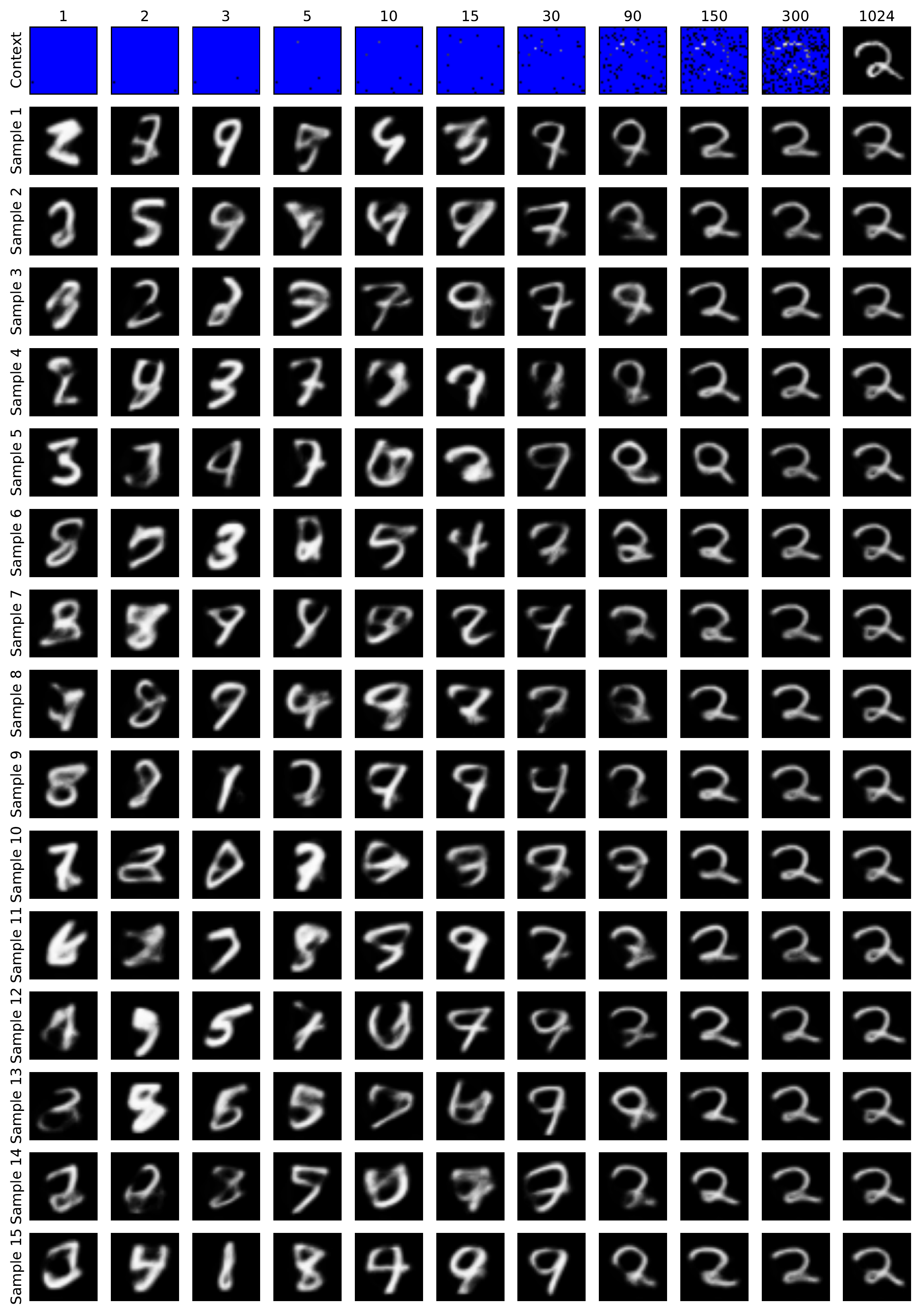}
        \caption{NP objective with max pooling}
    \end{subfigure}
    \hfill
    \begin{subfigure}[b]{0.45\linewidth}
        \centering
        \includegraphics[width=\linewidth]{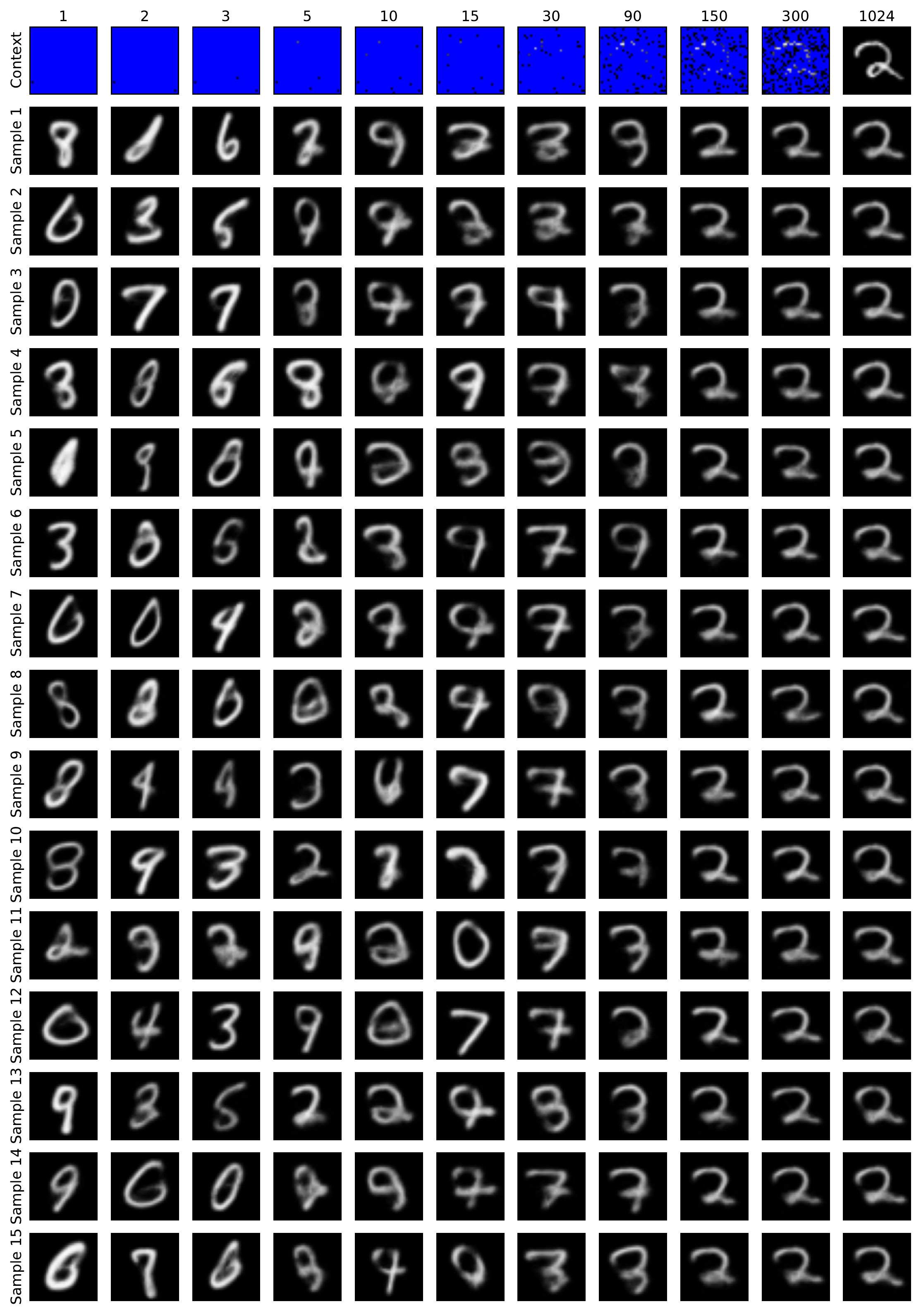}
        \caption{NP+SIVI objective with max pooling}
    \end{subfigure}
    \begin{subfigure}[b]{0.45\linewidth}
        \centering
        \includegraphics[width=\linewidth]{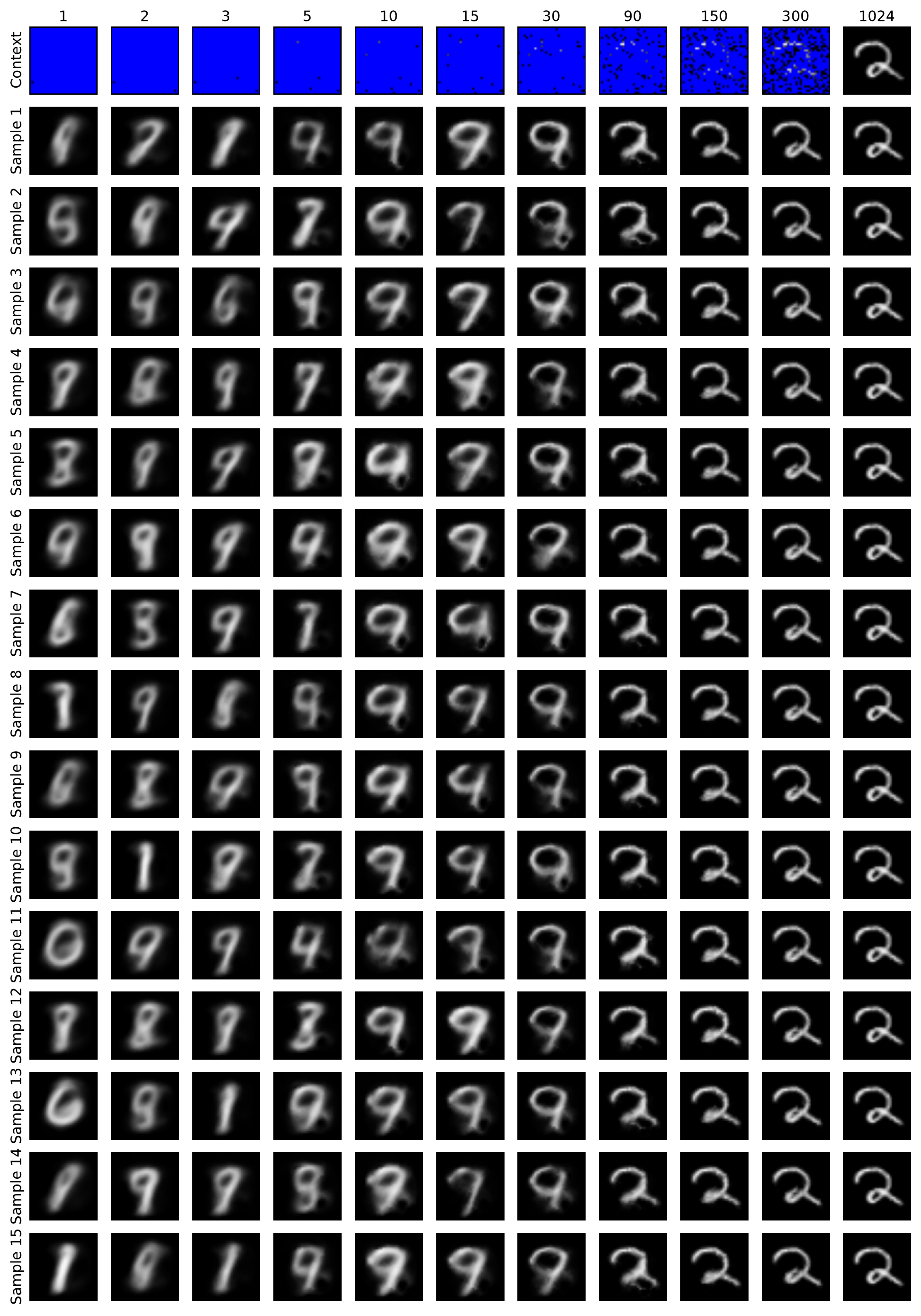}
        \caption{ANP}
    \end{subfigure}
    \caption{Samples from models trained on the MNIST dataset.}
    \label{app:additional-mnist-2}
\end{figure}

\begin{figure}[tbp]
    \centering
    \begin{subfigure}[b]{0.45\linewidth}
        \centering
        \includegraphics[width=\linewidth]{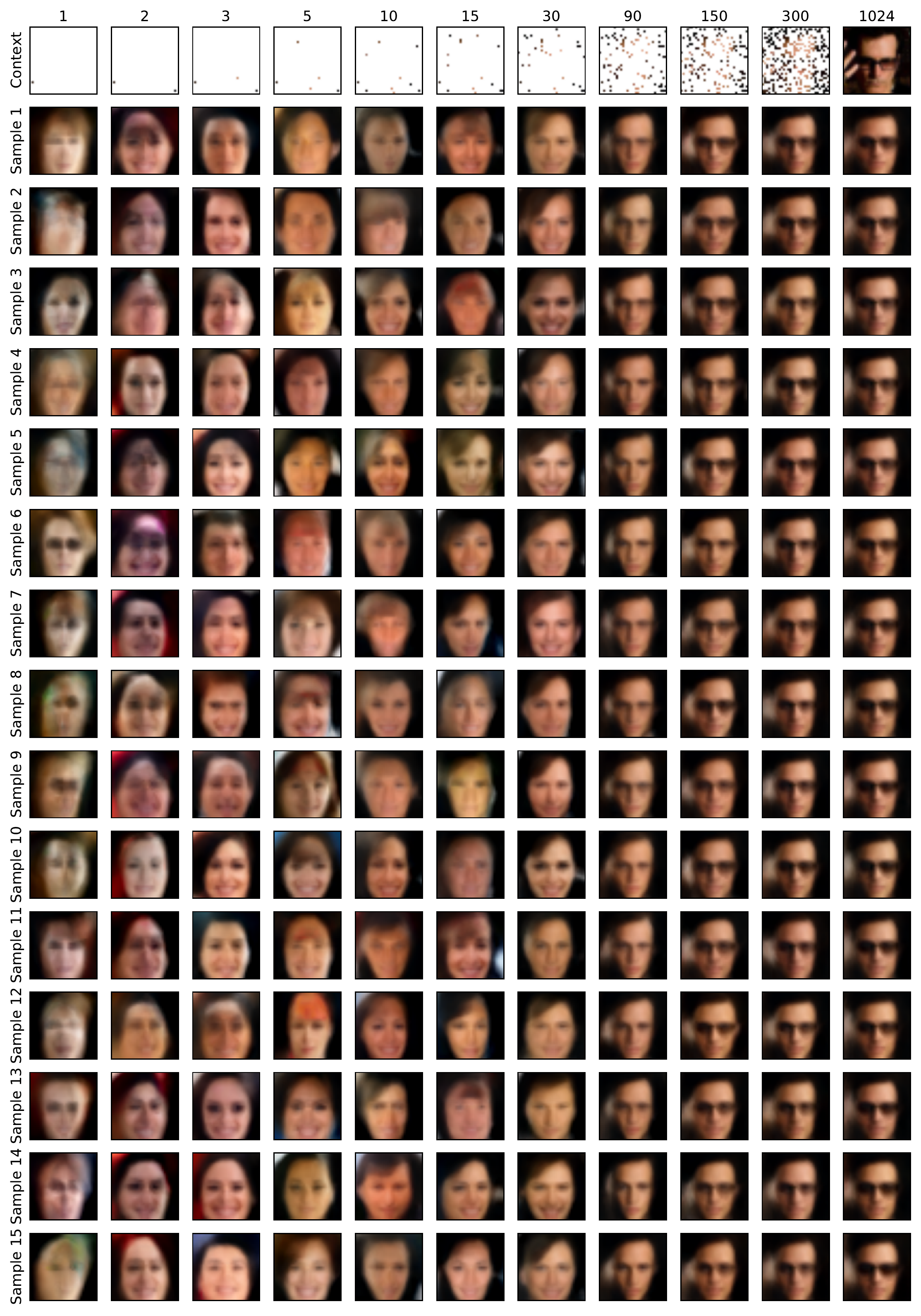}
        \caption{NP objective with average pooling}
    \end{subfigure}
    \begin{subfigure}[b]{0.45\linewidth}
        \centering
        \includegraphics[width=\linewidth]{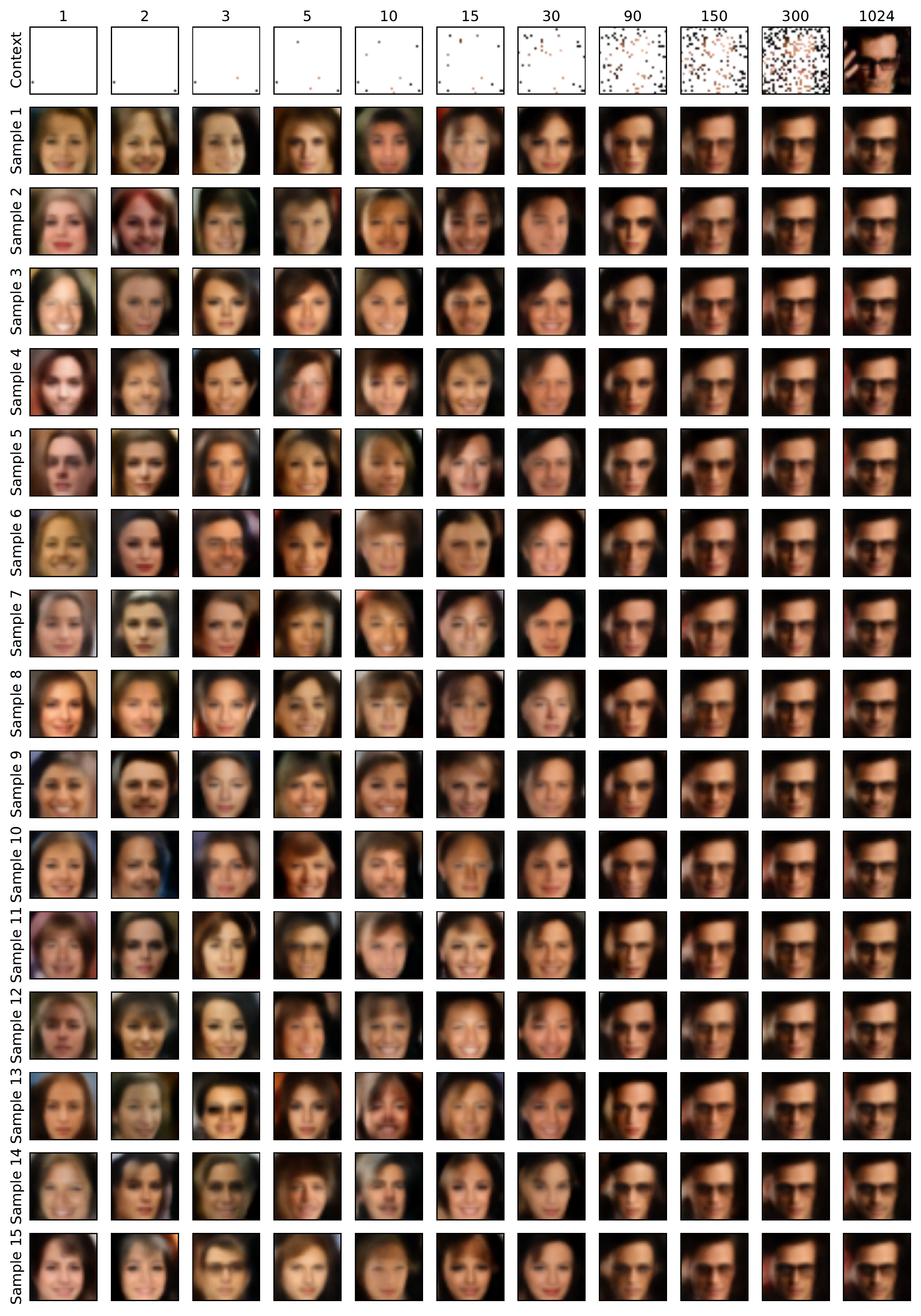}
        \caption{NP objective with max pooling}
    \end{subfigure}
    \hfill
    \begin{subfigure}[b]{0.45\linewidth}
        \centering
        \includegraphics[width=\linewidth]{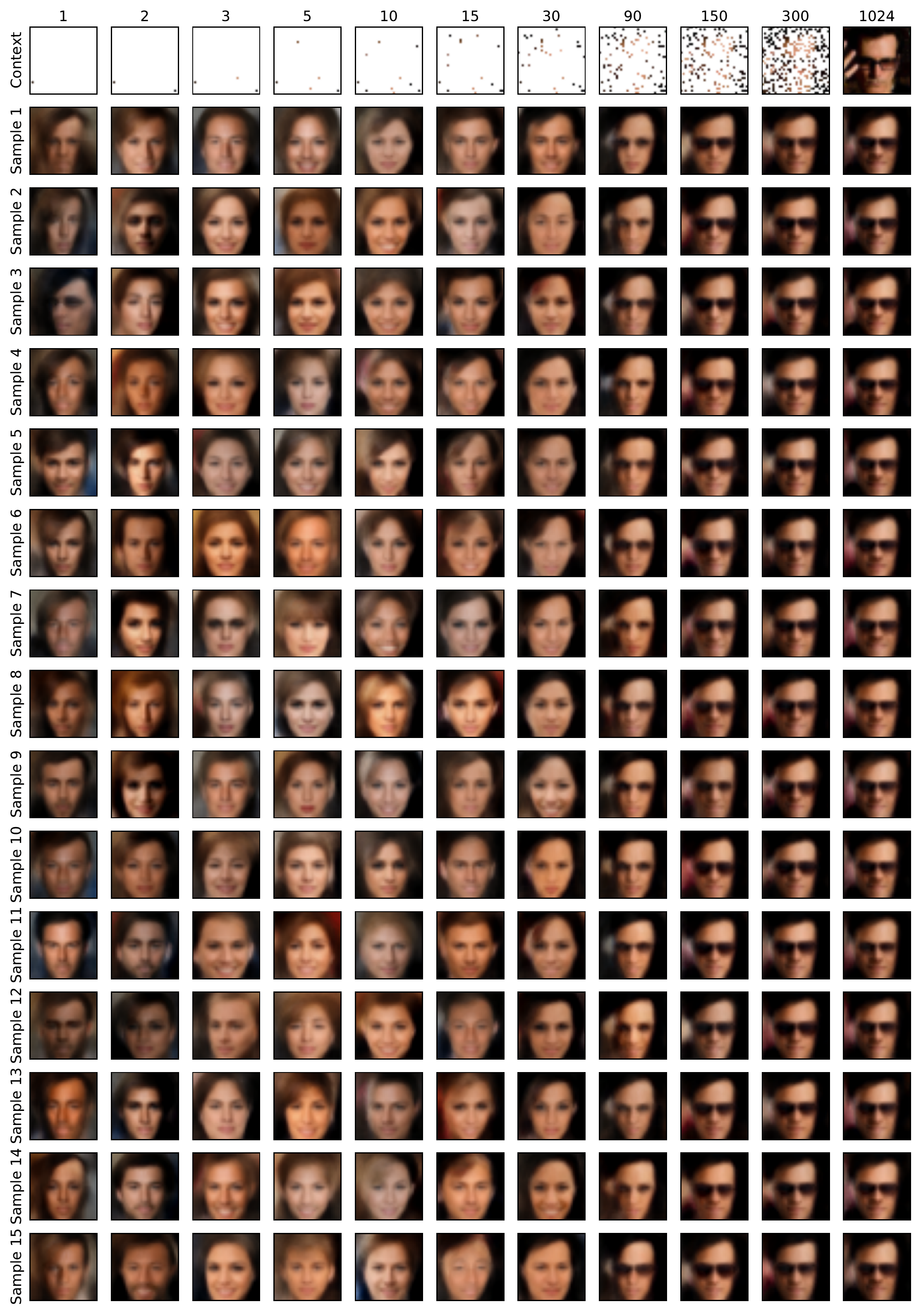}
        \caption{NP+SIVI objective with max pooling}
    \end{subfigure}
    \begin{subfigure}[b]{0.45\linewidth}
        \centering
        \includegraphics[width=\linewidth]{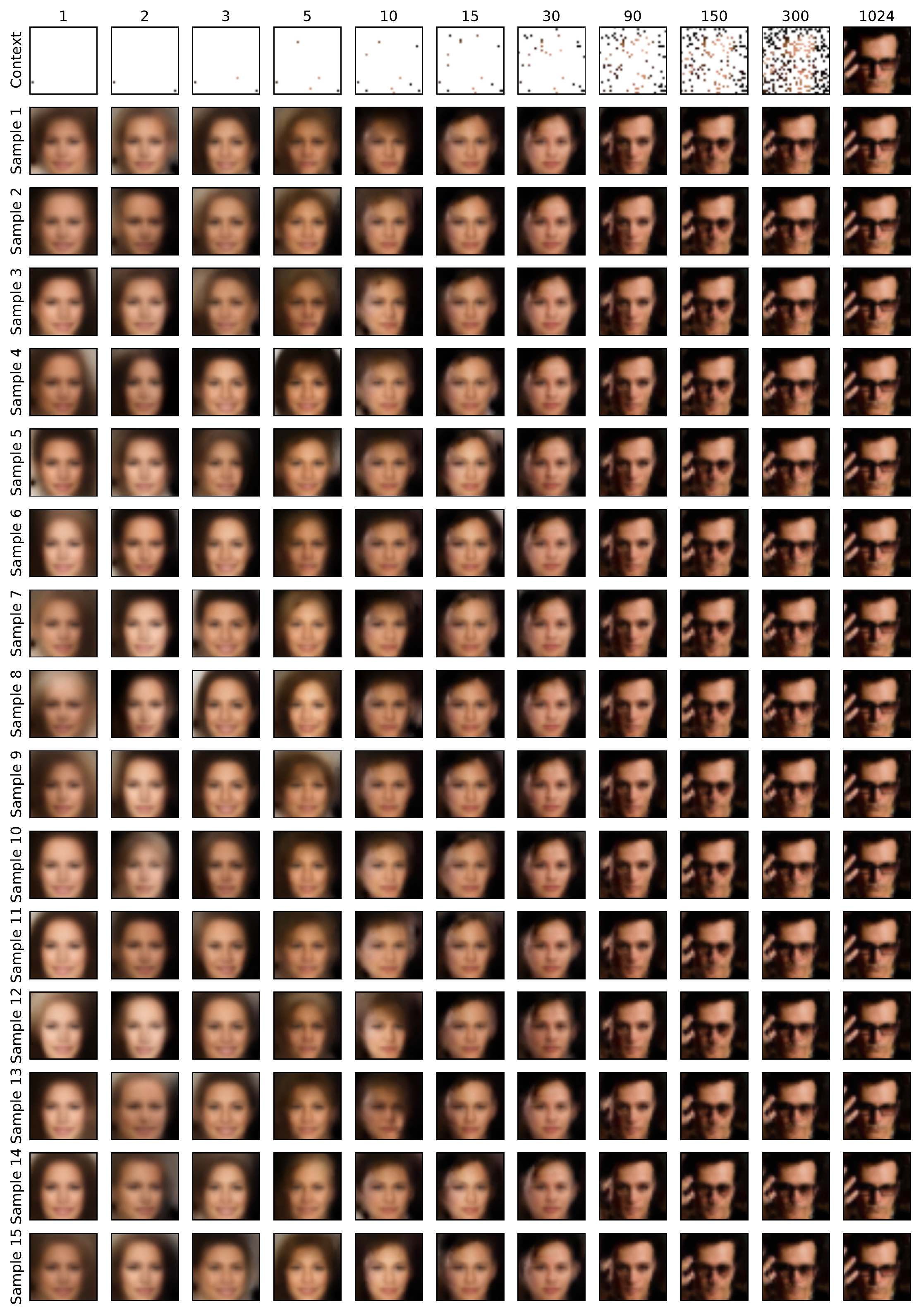}
        \caption{ANP}
    \end{subfigure}
    \caption{Samples from models trained on the CelebA dataset.}
    \label{app:additional-celeba-2}
\end{figure}

In this section, we report additional results for MNIST and CelebA experiments. \cref{app:additional-mnist,app:additional-mnist-2} show 15 samples per context set drawn from models trained on the MNIST dataset, and \cref{app:additional-celeba,app:additional-celeba-2} show the same for the CelebA dataset.


\end{document}